\documentclass[conference]{IEEEtran}
\IEEEoverridecommandlockouts

\usepackage{graphicx}
\usepackage{hyperref}

\usepackage{algorithm}
\usepackage[noend]{algpseudocode}
\usepackage{booktabs} 
\usepackage{multirow}

\usepackage{newfloat}
\usepackage{listings}
\usepackage{ntheorem}
\usepackage{amsmath}
\usepackage{amsfonts}
\usepackage{paralist}
\usepackage{dsfont}
\usepackage{caption}
\usepackage{subcaption}
\usepackage{xcolor}

\newtheorem{problem}{Problem}

\floatstyle{ruled}
\newfloat{listing}{tb}{lst}{}
\floatname{listing}{Listing}

\newcommand{\sstitle}[1]{\smallskip\noindent\textbf{#1.\/}}
\newcommand{\sititle}[1]{\smallskip\textit{#1:\/}}

\newcommand\Mark[1]{\textsuperscript{#1}}

\newtheorem{theo}{Theorem}
\newtheorem{defi}{Definition}

\author {
    Vinh Van Tong\Mark{1}, 
    Thanh Trung Huynh\Mark{2}, 
    Thanh Tam Nguyen\Mark{2},\\
    Hongzhi Yin\Mark{3}, 
    Quoc Viet Hung Nguyen\Mark{2}, 
    Quyet Thang Huynh\Mark{1}
\vspace{1.6mm}\\
\fontsize{9}{9}\selectfont\rmfamily\itshape
\small 
\Mark{1}Hanoi University of Science and Technology, Vietnam\\
\Mark{2}Griffith University, Australia\\
\Mark{3}The University of Queensland, Australia\\
}

\def\Snospace~{\S{}}

\begin{document}


\title{Incomplete Knowledge Graph Alignment}

\maketitle

\begin{abstract}

Knowledge graph (KG) alignment -- the task of recognizing entities referring to the same thing in different KGs -- is recognized as one of the most important operations in the field of KG construction and completion. 
However, existing alignment techniques often assume that the input KGs are complete and isomorphic, which is not true due to the real-world heterogeneity in domain, size, and sparsity. 
In this work, we address the problem of aligning incomplete KGs with representation learning. 
Our KG embedding framework exploits two feature channels: transitivity-based and proximity-based. The former captures the consistency constraints between entities via translation paths, while the latter captures the neighborhood structure of KGs via an attention guided relation-aware graph neural network. 
The two feature channels are jointly learned to exchange important features between the input KGs while enforcing the output representations of the input KGs in the same embedding space. 
Also, we develop a missing links detector that discovers and recovers the missing links in the input KGs during the training process, which helps mitigate the incompleteness issue and thus improve the compatibility of the learned representations. 
The embeddings then are fused to generate the alignment result, and the high-confidence matched node pairs are updated to the pre-aligned supervision data to improve the embeddings gradually. 
Empirical results show that our model is more accurate than the SOTA and is robust against different levels of incompleteness. 

 \end{abstract}


\section{Introduction}
\label{sec:intro}

Knowledge graphs (KGs) represent the facts about real-world entities in the form of triples $\langle head\_entity, relation, tail\_entity \rangle$ 
\cite{wang2017knowledge, sun2020benchmarking}. Popular knowledge graphs (e.g., DBpedia, YAGO, and BabelNet) are often multilingual, in which each language domain has a separate version~\cite{gracious2021neural}. To encourage the knowledge fusion between different domains, knowledge graph alignment -- the task of identifying entities in the cross-lingual KGs that refers to the same real-world object -- has received great interest from both industry and acadamia~\cite{trung2020adaptive, phan2018pair}. The alignment result can be used for further data enrichment applications such as repairing inconsistencies, filling knowledge gaps, and building cross-lingual KBs~\cite{wan2021gaussianpath,yan2021dynamic, zhang2018variational}.

\begin{figure*}[!ht]
	\centering
	\includegraphics[width=0.7\linewidth]{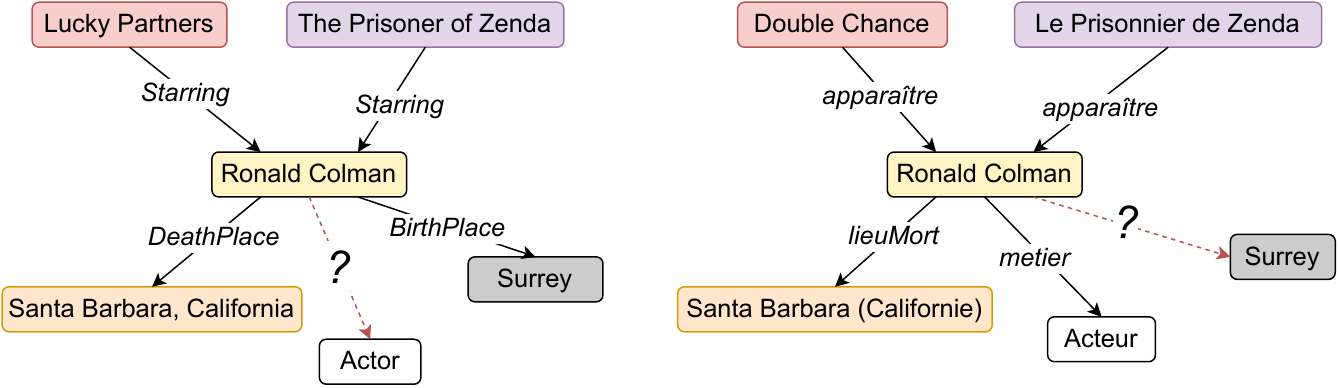}
    \caption{Aligning incomplete KGs across domains}
    \label{fig:example}
    \vspace{0em}
 \end{figure*}

Knowledge graph alignment faces challenges related to efficiency, scalability and richness of incorporated rich information of real-world KGs. Recent approaches solve these challenges by employing graph neural networks (GNNs) \cite{kipf2016semi, velivckovic2018graph} to encode the attributional and the relational triples via some message-passing schemes, e.g., GCN-Align~\cite{wang2018cross}, RDGCN~\cite{wu2019relation}, MUGNN~\cite{cao2019multi}, KG-matching~\cite{xu2019cross}. 

However, existing techniques often assume that the input KGs are nearly identical (isomorphic), which is not true~\cite{TRUNG2020112883, huynh2019network, sun2020knowledge}. There is often a considerable gap between the levels of completeness of different monolingual KGs \cite{sun2020benchmarking}, especially between the English domain and other languages \cite{zhao2020experimental}. For example, in the DBP15K dataset, the ratio of relational triples between Chinese, Japanese, or French KGs over the English KG is only around 80\% \cite{nguyen2020entity}. \autoref{fig:example} gives an example of incomplete KGs, in which the neighborhood of the two entities referring to the actor \textit{Ronald Colman} in English and French KGs are inconsistent (his occupation is missing in the English KG while his place of birth is missing in the French KG). Such inconsistencies easily lead to the different representations of the corresponding entities, especially for GNNs where the noise is accumulated over neural layers \cite{jumping_knowledge}.

In this paper, we address the above challenges by proposing a representation learning framework with \emph{multi-channel feature exchange} for aligning \emph{incomplete} knowledge graphs from {different domains}. To capture the multi-domain nature of KG entities, we develop a graph attention convolutional network that can combine the translated entity name (as the entity's feature) and the relational structure simultaneously. The attention mechanism allows relational importance integration, which helps mitigate the noise by focusing on mutual relations in the input KGs and ignoring the missing ones. Our proposed attention mechanism goes beyond the existing techniques \cite{wu2019relation, sun2020knowledge} by leveraging relation-aware attentive scoring, which helps the framework to integrate the KG edge information. To guarantee consistency across KGs, we develop an additional embedding component that encodes both entities and relations with the tradition translation constraint \cite{MTransE}. While many recent works neglect this `seems-to-be-strict' constraint~\cite{xu2019cross}, it turns out to strengthen the local information of relational triples and mitigate the information dilution phenomenon in graph convolutional networks~\cite{jumping_knowledge}. We also develop a missing links detector that consumes the two feature channels to exchange the knowledge from the two input KGs to discover and recover the missing triples. Finally, we combine these dimensions using late-fusion to instantiate the alignment result.

We summarise our contributions as follows:
\begin{itemize}
	\item We address the problem of aligning incomplete KGs from different domains and propose a framework called \textbf{I}ncomplete \textbf{K}nowledge graphs \textbf{A}ligner via \textbf{M}ult\textbf{I}-channel Feature Exchange (\textbf{IKAMI}). The model exploits multiple representations to capture the multi-channel nature of KGs (e.g., relational type, entity name, structural information). This is the first attempt to address the entity alignment and knowledge completion simultaneously, and we argue that this collaboration benefits both tasks, especially the alignment performance.

    \item We employ a translation-based embedding that encodes both entities and relations inspired from the translation model \cite{MTransE}. The translation assumption helps to align both entities and relations and is a great supplement to mitigate the information dilution weakness of the GCN-based embedding \cite{jumping_knowledge}.

    \item We propose a graph convolution attention network that efficiently captures relational triples, including the entity name, the relation type, and direction. The attention mechanism allows us to learn the underlying importance of relational triples based on their type, thus focusing on more popular relations and ignoring the ones that appear in only one network.

    \item We propose a missing triples detector that leverages the learned translation-based features to jointly discover and complete the missing links in the two input KGs. First, the potential missing links between the entities are proposed by selecting the ones with high-correlated embeddings, then the correct relation between them is chosen by selecting the appropriate relation embedding.  
    
    \item We design a joint train schedule of the two embedding models to enable the holistic objectives of the embeddings can support each other well. Then, the similarity matrix for each channel is calculated and fused by weighted-sum to return the final result.

    \item We conduct experiments on real-world and synthetic KG datasets to evaluate our scheme. The results show that our framework outperforms other baselines in not only the entity alignment task but also the knowledge completion task by up to 15.2\% and 3.5\%, respectively. We publish the source code for use by the community.

\end{itemize}

The remainder of the paper is organized as follows. \autoref{sec:related} reviews the related works and motivates our approach. \autoref{sec:background} presents a problem statement for the joint alignment and completion of incomplete knowledge graphs as well as our approach overview. \autoref{sec:model} describes the architecture of the two feature channels, including a translation-based embedding and a GCN embedding model that captures the relational correlation between the entities of KGs. \autoref{sec:process} explains how the current extracted feature channels are iteratively used to instantiate the alignment result and recover the missing triples from the input KGs, which helps to refine the alignment results gradually. Empirical evaluation is given in \autoref{sec:exp}, before \autoref{sec:con} presents a summary and conclusions.

\section{Related Work}
\label{sec:related}


\sstitle{Knowledge Graph Alignment}
Traditional cross-lingual KG alignment approaches often rely on various domain-independent features to discover cross-lingual links~\cite{el2015alex}. Modern cross-lingual KGs entity aligners exploit the emergence of graph embedding techniques and show promising results~\cite{chen2014unified, chen2019exploiting}. From the two input monolingual KGs, these techniques first embed the entities into low-dimensional vector spaces. The identical entities are then retrieved based on the similarity of the learned representations. The earlier techniques of this paradigm, including MTransE~\cite{MTransE}, JAPE~\cite{JAPE}, ITransE~\cite{zhu2017iterative}, and BootEA~\cite{sun2018bootstrapping}, often employ shallow embedding techniques to embed the entities and the relations using the translation constraint \cite{bordes2013translating}, which assumes that the embedding distance between two any entities is equal to the connecting relation.

Due to the lack of information integration of shallow models, recent embedding based alignment techniques employ graph convolutional network (GCN) ~\cite{kipf2016semi} to exploit the complex nature of KGs better. GCN-Align~\cite{wang2018cross} employs GCNs to capture at the same time the relational and the individual entity footprint. RDGCN~\cite{wu2019relation} introduces a two-layers GCN with highway gates to extract attentive interactions between each KG and its dual relation. MUGNN~\cite{cao2019multi} preprocesses the input KGs with grounding rules before forwarding them to a dual-channel GCNs model. KDcoE \cite{chen2018co} leverages co-training of a KG embedding and a literal description embedding to enhance the semi-supervised learning of multilingual KG embeddings. MultiKE \cite{zhang2019multi} unifies multiple views of entities to learn embeddings for entity alignment. KG-matching~\cite{xu2019cross} constructs a relation network namely \textit{topic entity graph} to encode the entities' name by a word-based LSTM. REA \cite{pei2020rea} proposes a two-component model to enhance the noise robustness of the aligner. AliNet \cite{sun2020knowledge} employs entities' distant neighbours to expand their neighbour overlapping~\cite{chen2021passleaf}.

Our work goes beyond the state-of-the-art by enhancing the GNN properties with attention mechanism and entity-relation composition to capture the relational correlation effectively. We also combine the GNN with a translation-based channel to guarantee the knowledge transfer between the input KGs and thus mitigate the incompleteness issue. Similar to our work are GCN-Align and MultiKE, which also use a multi-view scheme, but our GNN channel leverages relation embeddings that better capture relational information. Also, unlike the existing models such as RDGCN and GCN-Align that stacks GNN models, our framework utilizes a non-GNN channel to cover the GNN-based channel and overcome the GNN weaknesses, such as information dilution and noise amplification issue.

\sstitle{Knowledge Graph Completion} The Knowledge Graph Completion (KGC) techniques aim to predict the missing triples in the KGs automatically. Existing KGC techniques often embed entities and relations into low-dimensional vectors, then design a score function to check the candidate triplets' plausibility. The earlier methods often employ shallow embedding models with translation-based scoring function \cite{bordes2013translating, wang2014knowledge}. Recently, KGC techniques using deep embedding model and complex score have been proposed, including CNNs based model \cite{dettmers2018convolutional, nguyen2018novel}, RNNs based models \cite{liu2017analogical}, and GCNs based model \cite{shang2019end}. However, they are surprisingly found to be unstable across different datasets due to the overloading of unrelated objectives \cite{sun2020re}. Thus, we choose the simple yet powerful translation-based model to supply the main GNN-based channel in our framework. Our work also goes beyond the existing KGC framework by answering at the same time two questions: (i) given any two entities, whether there is a triple connecting them and (ii) which is the relation type of the triple between the two entities; while the existing frameworks often answer only the latter question. It is worth noting that the former question is challenging, as the number of disjoint entities outweighs that of connecting ones and thus might cause low recall in detecting the missing triples. We tackle this challenge by exchanging the useful patterns of high-confidence entities retrieved from the alignment of the two input KGs, thus making the information of the KGs complete for each other.

\section{Incomplete Knowledge Graph Alignment}
\label{sec:background}

In this section, we formulate the problem and discuss the challenges for incomplete knowledge graph alignment as well as our approach overview.

\subsection{Problem Formulation}

\sstitle{Incomplete knowledge graphs (\emph{i}-$\mathcal{KG}s$)} The KG is often denoted as $\mathcal{KG} = (\mathcal{V}, \mathcal{R}, \mathcal{E})$, where $\mathcal{V}$ is the set of entities; $\mathcal{R}$ is the set of relations and $\mathcal{E}$ is the set of triples. The triple $\langle h, r, t \rangle \in \mathcal{E}$ is atomic unit in $\mathcal{KG}$, which depicts a relation \textit{r} between a head (an entity) \textit{h} and a tail \textit{t} (an attribute or another entity). We present the incomplete knowledge graphs by extending the $\mathcal{KG}$ notation as \emph{i}-$\mathcal{KG} = (\mathcal{V}, \mathcal{R}, \mathcal{E}, \bar{\mathcal{E}})$, where $\bar{\mathcal{E}}$ is the set of missing triples in the \emph{i}-$\mathcal{KG}$. For brevity's sake, we use \emph{i}-$\mathcal{KG}$ and $\mathcal{KG}$ interchangeably in this paper.

\sstitle{Incomplete knowledge graph alignment}
By generalising the problem setting in related works, \emph{i}-$\mathcal{KG}$ alignment aims to find all of the corresponding entities of two given \emph{i}-$\mathcal{KG}s$. Without loss of generality, we select one \emph{i}-$\mathcal{KG}$ as the source graph and the other as the target graph, and denote them as $\mathcal{KG}_s = (\mathcal{V}_s, \mathcal{R}_s, \mathcal{E}_s, \bar{\mathcal{E}}_s)$ and $\mathcal{KG}_t = (\mathcal{V}_t, \mathcal{R}_t, \mathcal{E}_t, \bar{\mathcal{E}}_t)$ respectively. Note that $\mathcal{E}_s \bigcup \bar{\mathcal{E}}_s = \mathcal{E}_t \bigcup \bar{\mathcal{E}}_t$, which represents the complete triple facts.  Then, for each entity $p$ in the source graph $\mathcal{KG}_s$, we aim to recognise its counterpart $p'$ (if any) in the target knowledge graph $\mathcal{KG}_t$. The corresponding entities $(p, p')$ also often denoted as anchor links; and existing alignment frameworks often require supervision data in the form of a set of pre-aligned anchor links, denoted by $\mathbb{L}$.

Since the corresponding entities reflect the same real-world entity (e.g. a person or concept), the existing alignment techniques often rely on the \textit{consistencies}, which states that the corresponding entities should maintain similar characteristics across different $\mathcal{KG}$s \cite{wang2018cross}. The \textit{entity consistency} states that the entities referring to the same objects should exist in both the KGs and have equivalent name. The relation consistency (a.k.a. the \emph{homophily rule}) declares that the entities should maintain their relationship characteristics (existence, type, direction). 


\sstitle{Knowledge graph completion} Given the incomplete knowledge graph \emph{i}-$\mathcal{KG} = (\mathcal{V}, \mathcal{R}, \mathcal{E}, \bar{\mathcal{E}})$, where $\bar{\mathcal{E}}$ is unrevealed, the knowledge graph completion (KGC) task aims to discover all the missing triples $ \langle h, r, t \rangle \in \bar{\mathcal{E}} | \langle h, r, t \rangle \notin \mathcal{E}$.

While KG alignment and completion have been studied for decades~\cite{sun2020knowledge,sun2020re}, there is little work on jointly solving these problems together. However, doing so is indeed beneficial: missing triples $ \langle h, r, t \rangle \in \bar{\mathcal{E}}$ in one $\mathcal{KG}$ can be recovered by cross-checking another $\mathcal{KG}$ via the alignment, which, in turn, can be boosted by recovered links. To the best of our knowledge, this work is a first attempt to solve the joint optimization of KG alignment and completion, which is formally defined as follows.

\begin{problem}[Joint KG alignment and completion]
\label{prob:joint}
Given two incomplete knowledge graphs $\mathcal{KG}_s$ and $\mathcal{KG}_t$, the task of joint KG alignment and completion is to: (i) identify all the hidden anchor links between $\mathcal{KG}_s$ and $\mathcal{KG}_t$, and (ii) recover the missing triples in each input  $\mathcal{KG}_s$ and $\mathcal{KG}_t$. 
\end{problem}

\autoref{tab:notation} summarizes the important notations used in this paper.

\begin{table}[!h]
\vspace{-5pt}
\centering
\caption{Summary of notation used}     \label{tab:notation}
\scriptsize
\resizebox{\linewidth}{!}{%
\begin{tabular}{c l}
\toprule
\textbf{Symbols} & \textbf{Definition} \\
\midrule
$\mathcal{KG} = (\mathcal{V}, \mathcal{R}, \mathcal{E}, \bar{\mathcal{E}})$ & Incomplete Knowledge Graph \\
$p$, $q$, $h$, $t$ & Entity \\
    $r$ & Relation \\
    $\mathbf{h}_p$, $\mathbf{h}_r$ & Entity embedding \& Relation embedding \\
    $\mathcal{V}$, $\mathcal{R}$, $\mathcal{E}$, $\bar{\mathcal{E}}$   & Entity set, Relation set, Triple set, Missing triple set  \\
    $\mathcal{N}(p)$ & Neighbor set of node $p$ \\
\midrule
$K$ & Number of GNN layers \\
$d$ & Embedding dimension \\
$\gamma_t$ & The margin of translation loss in \textit{transitivity-based channel}\\
$\gamma_g$ & The margin of entity alignment loss in \textit{proximity-based channel} \\
$\beta_t$ & Balancing weight for loss terms in \textit{transitivity-based channel} \\
$\beta_g$ & Balancing weight for loss terms in \textit{proximity-based channel} \\
$\beta_{tg}$ & Balancing weight for alignment matrices of the two channels \\
\midrule
$\mathbf{W}_{\lambda(r)}$ & Direction-aware projection matrix \\
$\mathbf{W}_r^k$ & Relation transformation matrix at k-th layer \\
$\mathbf{W}_{att}^k$ & Attention weight matrix at k-th layer \\
$\mathbf{W}_{ge}$ & GNN's final projection matrix for entities \\
$\mathbf{W}_{gr}$ & GNN's final projection matrix for relations \\
\midrule
$\mathbf{S}$ & Alignment matrix \\ 

\bottomrule
\end{tabular}
}
\vspace{-5pt}
\end{table}

\subsection{Challenges}

Solving \autoref{prob:joint} is non-trivial. We argue that an efficient incomplete $\mathcal{KG}$ entity alignment framework should overcome the following challenges:

\begin{enumerate}[C1.]
    \item \textit{Domain gap:} As the input $\mathcal{KG}$s are incomplete, there exist inconsistencies between the cross-lingual $\mathcal{KG}$s (i.e. incompatible individual information, inequivalent neighbor set, different number of nodes in each graph). Existing works attempt to tackle this challenge by applying rule-base $\mathcal{KG}$ completion as preprocessing step \cite{cao2019multi, pei2020rea}, but this fails to leverage the correlation between the two input $\mathcal{KG}$s and requires the addition of pre-aligned relation types information. 
    
    \item \textit{Task gap:} While the incompleteness might cause the inconsistencies, the consistencies (entity consistency, relational consistency) should be respected overall since these constraints guide to finding precisely matched entities w.r.t the $\mathcal{KG}$s specific characteristics (e.g., name equivalence, directional relations). 
Therefore, handling the two ``seem-to-contradict" tasks, KG completion and KG alignment, simultaneously is challenging. 
    
    
    \item \textit{Model gap:} The neighbourhood structures of $\mathcal{KG}$s provide rich information in various forms (e.g. relational triples, relational directions). Recent works exploit this characteristic by stacking GCNs in their model \cite{wang2018cross, wu2019relation}, but the structure of the GCNs are often similar and thus suffer from the same weakness. 

\end{enumerate}

\subsection{Outline of the Alignment Process}

\begin{figure*}[t]
\centering
\includegraphics[width=\linewidth]{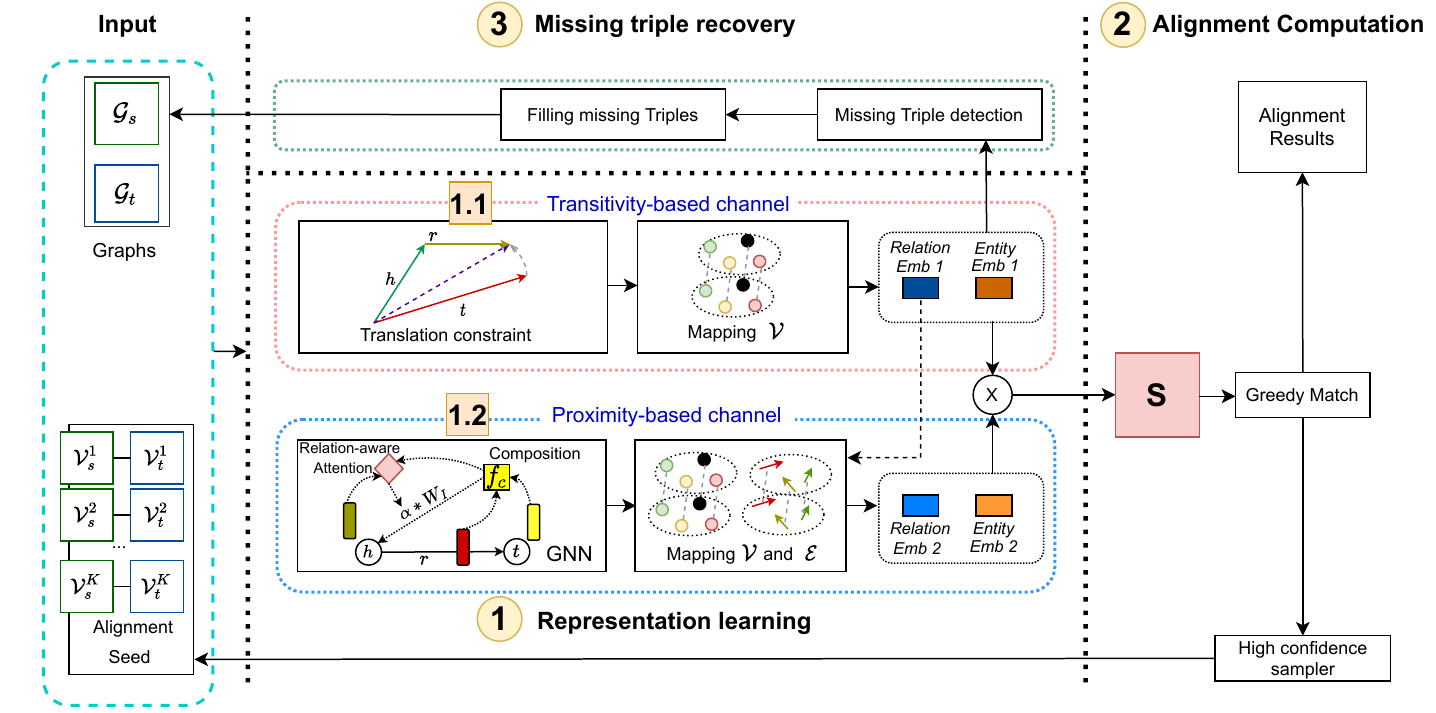}
    \caption{Framework Overview}
    \label{fig:framework}
    \vspace{0em}
\end{figure*}

To address the above challenges, we argue that an alignment of incomplete knowledge graphs shall happen iteratively. \autoref{fig:framework} shows an overview of our alignment process. In each step, the input KGs are enriched (missing triples are recovered), improving their alignment simultaneously. Such incremental process allows the hard cases to be more likely solved as the model experiences easier cases and becomes more mature over time.

Starting with two incomplete KGs $\mathcal{KG}_s$ and $\mathcal{KG}_t$, the alignment process continuously updates three objects:
\begin{compactitem}
	\item \emph{Alignment matrix} $\mathbf{S} \in \mathbb{R}^{|\mathcal{V}_s| \times |\mathcal{V}_t|}$ that represents the result of alignment between the source and target graphs, where each $|\mathcal{V}_s|$ and $|\mathcal{V}_t|$ denotes the numbers of entities in $\mathcal{KG}_s$ and $\mathcal{KG}_t$, respectively. Each component $\mathbf{S}(p,p')$ in the alignment matrix $\mathbf{S}$ identifies the alignment level between an entity $p \in \mathcal{V}_s$ and its counterpart entity $p' \in \mathcal{V}_t$.
	\item \emph{Alignment seed} $\mathbb{L}$ that is a set of known aligned entities.
	\item \emph{Missing triples} $\bar{\mathcal{E}}$ of the KGs themselves.
\end{compactitem}

Each iteration of the alignment process comprises the following steps:

\begin{enumerate}[(1)]
	\item[(1)] \emph{Representation learning:} We first forward the source and target KG networks through two designed feature channels, namely \textit{transitivity-based channel} and \textit{proximity-based channel}, to embed the KG entities in different low-dimensional vector spaces. These two channels are parallel:
	\item[(1.1)] \emph{Proximity-based channel:} is a GCN-based model that unifies the entity name information and topological structure under the same modal. To this end, we overcome the language barrier (C1) by employing the word embedding of the translated entity name. The channel also produce the relation embedding as well as utilize an attention mechanism to exploit the relational information, which helps to detect the proximity of the corresponding entities in many aspects and thus helps to narrow down the language gap (C1) and mitigate the possible noises (C2). The details of this step can be found in \autoref{sec:transitivity}.
	
	\item[(1.2)] \emph{Transitivity-based channel} is a shallow-based embedding model that enforces the translation constraint between the embeddings of entities and relations in each triple. This constraint helps to emphasize on the topological first-order proximity information, which thus helps to cover the information dilution issue of GNN based model (C3). Also, the translation constraint encourages the analogy between the source and target KGs embedding space (as the relation type between corresponding entities reflect the same phenomenon), which reduces the noise (C2) and facilitates the feature exchange using the learnt embedding spaces (C1). We discuss about this channel in \autoref{sec:proximity}.
	
	\item[(2)] \emph{Alignment computation:} The learned representations are then fused to compute the final alignment result. We first train the Transitivity-based channel to get the representation of entities and relations. Then we train the Proximity-based channel using the input graphs structure as well as entity name embedding. The relation representations of the Transitivity-based channel are also used as input to this channel to allow the relation-aware mechanism of the GNN to work properly. The output of this channel are relation and entity embeddings. The embedding of entities of the two channels then are concatenated and applied a linear transformation to get the final embedding. We then compute the cosine similarity between each pair of entities across two $\mathcal{KG}$s to get the alignment matrix. The high-confidence matched entities then is sampled and added to the alignment seed $\mathbb{L}$, which is used to enhance the quality of the two representation channels in the next iterations. The detailed implementation of this step can be found in \autoref{sec:alignment}.
	
	\item[(3)] \emph{Missing triples recovery:} We develop a two-step module that leverages the learnt representations are used to recover all possible missing triples in the input KGs. At the first step, a 2-layer perceptron followed by a sigmoid function is used to compute the probability of whether there exists a missing relation between the two entities. If the answer is yes, indicated by the probability being above a pre-defined threshold, we determine which type of relation connecting the entities in the second step. In more details, given the two entities $p, q$ and their \textit{transitivity-based} embeddings, we choose the relation $r$ whose embedding follows the translation constraint $p + r \approx q$. Note that we do not use directly the embeddings of the \textit{proximity-based channel} due to the incapability of GNN-based model in capturing positional information \cite{jumping_knowledge}; but this important channel plays an important role in finding hidden anchor links and update the \textit{transitivity-based} channel. Further details of this process is presented in \ref{sec:recovery}
	
\end{enumerate}

The alignment process is stopped when the validation accuracy can not further increase after several consecutive iterations. The final alignment matrix is then used to retrieve the matched entities.

\section{Feature channel models}
\label{sec:model}

\subsection{Pre-processing} 

To efficiently capture the $\mathcal{KG}$ relation direction, we preprocess the input $\mathcal{KG}$s by adding inverse triples and self-loop triples as follows: 
\begin{equation}
    \label{eqn:prep_entity}
 \mathcal{E} = \mathcal{E} \cup \{\langle h, r^{-1}, t \rangle | \langle h, r, t \rangle \in \mathcal{E}\} \cup \{\langle p, \top, p \rangle | p \in \mathcal{V} \} 
 \end{equation}

\noindent
and
 \begin{equation}
    \label{eqn:prep_rel}
  \mathcal{R} = \mathcal{R} \cup \mathcal{R}_{inv} \cup \{\top\}
\end{equation}

\noindent
where $\mathcal{R}_{inv} = \{r^{-1}| r \in \mathcal{R}\}$ represents the inverse relations and $\top$ denotes the self loop. The inverse edges helps the information can freely propagate in both direction in the learning step, while the adding of the self loop relation enables our GNN to passing message from one node to it-self.

\subsection{Transitivity-based channel}
\label{sec:transitivity}


The main role of this channel is to embed both entities and relations of the two $\mathcal{KG}$s to a same vector space so that those presentations can preserve the structure of the two graphs. To do that, we make use of a well-known translation constraint \cite{bordes2013translating} to all of the triples of the two graphs. We also apply a mapping loss on entities in the alignment seed to make sure the embeddings of the two graphs are in the same vector space. We also find out that by doing this, not only entities but also relations of the two graphs are aligned.

Formally, we employ a shallow translation-based embedding as an additional channel to complete the ``deep" GNN-based embedding. To this end, for each entity $p$ and relation $r$, we assign a trainable representation vector $\mathbf{h}_{(.)}$. The backbone of the model is the translation constraint \cite{bordes2013translating}, which enforces that for any relational triples $\langle h,r,t \rangle \in \mathcal{E}$, the following constraint should hold:
\begin{equation}
    \mathbf{h}_h + \mathbf{h}_r = \mathbf{h}_t
    \label{eqn:trans_constraint_ori}
\end{equation}
where $\mathbf{h}_h$, $\mathbf{h}_r$ and $\mathbf{h}_t$ are the embedding of the head, relation and tail entity of the triple, respectively. To integrate this constraint into the model, we employ the dissimilarity measure $d_t$ and guarantee that $d_t(h + r, t) \approx 0$ if $\langle h, r, t \rangle \in \mathcal{E}$, otherwise $d_t(h + r, t) >> 0$. In our work, we choose $d_t$ as Manhattan distance, similar to \cite{MTransE}.

\sititle{Translation loss} Given the translation dissimilarity function $d$ and the supervised triples set $\mathcal{E}$, we optimize the translation loss as follows:
\begin{equation}
    \mathcal{L}_{ts} = \sum_{(h,r,t) \in \mathcal{E}^+}{\sum_{(h',r,t') \in \mathcal{E}'}{[\gamma_t + d_t(h + r,t) - d_t(h' + r, t')]_+}}
    \label{eqn:trans_loss}
\end{equation}
where the $[x]_+$ denotes the positive part of $x$, $\gamma_t$ is a margin hyper-parameter, $\mathcal{E}'$ is the negative triple set constructed by corrupting either the head or tail in the original triples.


\sititle{Mapping loss} To jointly embed both graphs into the same embedding space, we will minimize the distance between entity pairs in the alignment seeds $\mathbb{L}$:
\begin{equation}
    \mathcal{L}_{tm} = \sum_{(p, p') \in \mathbb{L}} d_t\left(p, p'\right)
\end{equation}
The final loss function thus has the form:
\begin{equation}
    \mathcal{L}_{t} = \mathcal{L}_{ts} +  \beta_{tm} \mathcal{L}_{tm}
\label{eqn:loss1}
\end{equation}
where $\beta_{tm} \in \mathbb{R}$ is a hyper-parameter that scales the important of $\mathcal{L}_{tm}$. Note that the loss function $\mathcal{L}_{t}$ (\autoref{eqn:loss1}) allows the model to align relations without knowing any pre-aligned relation seeds. This is because the transitivity constraint can guarantee that if the corresponding entities in the input KGs are embedded into a space with the same representations, the same goes with the relation embeddings. The following theorem support this argument:  



\begin{defi}
($\epsilon$-closed entity pair): Given a real number $\epsilon \geq 0$ and two entities $p \in \mathcal{V}_1$ and $q \in \mathcal{V}_2$; an entity pair $(p, q)$ across $\mathcal{KG}$s is called an \textit{$\epsilon$-closed entity pair} if $d_t(p, q) \leq \epsilon$.
\end{defi}

\begin{defi}
($\epsilon$-closed relation pair): Given a real number $\epsilon \geq 0$ and two relations $r \in \mathcal{E}_1$ and $r' \in \mathcal{E}_2$; a relation pair $(r, r')$ across $\mathcal{KG}$s is called an \textit{$\epsilon$-closed relation pair} if $d_t(r, r') \leq \epsilon$.
\end{defi}

\begin{theo}
\label{theo:rel_align}
If $(p, q)$ and $(p', q')$ are two  \textit{$\epsilon$-closed entity pair}s, which means $d_t(p, q) \leq \epsilon$ and $d_t(p', q') \leq \epsilon$, then for any $r_{pq}$ connecting $p$ to $q$ and $r_{p'q'}$ connecting $p'$ to $q'$ such that the translation constraint in \autoref{eqn:trans_constraint_ori} is satisfied, then $(r_{pq}, r_{p'q'})$ is a $2\epsilon$-\textit{closed relation pair}. 
\end{theo}

\textit{Proof:} Suppose $\mathbf{h}_p$, $\mathbf{h}_q$, $\mathbf{h}_{p'}$, $\mathbf{h}_{q'}$, $\mathbf{h}_{r_{pq}}$, and $\mathbf{h}_{r_{p'q'}}$ are embeddings of $p$, $q$, $p'$, $q'$, $r_{pq}$, and $r_{p'q'}$ respectively. We have the distance between $r_{pq}$ and $r_{p'q'}$ is:
\begin{equation}
    \nonumber
    d_t(r_{pq}, r_{p'q'}) = ||\mathbf{h}_{r_{pq}} - \mathbf{h}_{r_{p'q'}}||_{L_1}
\end{equation}
As \autoref{eqn:trans_constraint_ori} is satisfied, we can replace $\mathbf{h}_{r_{pq}}$ by $(\mathbf{h}_q - \mathbf{h}_p)$ and $\mathbf{h}_{r_{p'q'}}$ by $(\mathbf{h}_{q'} - \mathbf{h}_{p'})$ to achieve:
\begin{equation}
\nonumber
\begin{split}
d_t(r_{pq}, r_{p'q'}) = ||(\mathbf{h}_q - \mathbf{h}_p) - (\mathbf{h}_{q'} - \mathbf{h}_{p'})||_{L_1} \\ 
= ||(\mathbf{h}_q - \mathbf{h}_p) + (\mathbf{h}_{p'} - \mathbf{h}_{q'})||_{L_1} \\
\leq ||\mathbf{h}_q - \mathbf{h}_p||_{L_1} + ||\mathbf{h}_{p'} - \mathbf{h}_{q'}||_{L_1} \\
= d_t(p, q) + d_t(p', q') \\
\leq \epsilon + \epsilon = 2 \epsilon   
\end{split}
\end{equation}

Thus, we can conclude that $(r_{pq}, r_{p'q'})$ is a \textit{$2\epsilon$-closed relation pair}. This valuable characteristic would also be used in the next \textit{proximity-based channel} to make sure our relation-aware mechanism work.

\subsection{Proximity-based channel} 
\label{sec:proximity}

In this channel, we take both entity name information and the local neighborhood structure around each entities into considerations. To better alleviate the relation information, we design a specific GNN architecture that allows our model to be relation aware. This GNN has two main innovations namely relation-aware message and relation-aware attention. We apply the mapping loss function on entities alignment seeds. We also introduce a new loss component to map the relation embeddings to the same vector space using the relation representations in the Transitivity-based channel.

Formally, our proximity-based channel is designed to unify $\mathcal{KG}$s heterogeneous information under the same modal using deep neural network model consisting of $K$ layers. For the $k$-th layer, we update the representation $\mathbf{h}_p^{k + 1} \in \mathbb{R}^d$ of each entity $p \in \mathcal{V}$ by:
\begin{equation}
    \mathbf{h}_p^{k + 1} = f_e \left( \sum_{(q, r) \in \mathcal{N}(p)} \alpha_{pqr}^k \mathbf{m}^k_{qr} \right) 
    \label{eqn:entity_update}
\end{equation}
where $\mathcal{N}(p) = \left\{(q, r) | (\langle p, r, q \rangle \in \mathcal{E}) \vee (\langle q, r, p \rangle \in \mathcal{E}) \right\}$ is the neighbor set of entity $p$, $\mathbf{m}^k_{qr} \in \mathbb{R}^d$ denotes the message passing from neighbor entity $p$ to entity $q$ through relation $r$, $\alpha^k_{pqr}$ represents the attention weight that emphasize the importance of the relational message $\mathbf{m}^k_{qr}$ to $p$; and $f_e$ is a linear transformation followed by a $Tanh(.)$ activation function. The innovations of our GNN-based model are two-fold: (i) we guarantee the message-passing is relation-aware by learning the relation embedding $\textbf{h}^{k}_r$ for each relation $r \in \mathcal{R}$ and integrating relation semantic into the entity message propagation $\mathbf{m}^k_{qr}$ and (ii) we design the attention weight $\alpha^k_{pqr}$ that further enhance the relation-aware capability of our GNN-based embedding.

\sititle{Relation-aware message} Unlike existing GNN-based techniques that often infer relation embeddings from learnt entity embeddings \cite{sun2020knowledge}, our technique allows entity and relation embeddings to be learnt independently and jointly contribute to the message passing process by applying the entity-relation composition operations:
\begin{equation}
    \mathbf{c}^k_{qr} = f_c \left(\mathbf{h}^k_q, \mathbf{h}^k_r\right)
\end{equation}
where $f_c: \mathbb{R}^d \times \mathbb{R}^d \rightarrow \mathbb{R} ^ d$ is a composition operator, and $\mathbf{c}^k_{qr} \in \mathbb{R}^d$ is the composition output vector. We choose the composition operator as substraction function \cite{bordes2013translating} given its simplicity, non-parameter and efficiency. We then compute the message from $q$ to $p$ through $r$ by projecting the composition output based on its direction:
\begin{equation}
    \mathbf{m}_{qr}^k = \mathbf{W}_{\lambda(r)} \mathbf{c}^k_{qr}
\end{equation}
\begin{equation}
    \nonumber
    \mathbf{W}_{\lambda(r)} = \left\{ \begin{array}{ll}
          \mathbf{W}_I, & \text{if } r \in \mathcal{R}_{inv} \\
          \mathbf{W}_S, & \text{if } r = \top \text{(self-loop)} \\
          \mathbf{W}_O, & \text{otherwise}
    \end{array}\right.
\end{equation}
Then, along with the entity embdding update in \autoref{eqn:entity_update}, the relation embedding is updated by:
\begin{equation}
    \mathbf{h}_r^{k + 1} = \mathbf{W}_{r}^k \mathbf{h}_r^k
    \label{eqn:relation_update}
\end{equation}

\noindent
where $\mathbf{W}_{r}^k \in \mathbb{R}^{d \times d}$ is a trainable transformation matrix that projects all the relations to the same embedding space and allows them to be utilized in the next layer. 


\sititle{Relation-aware attention} Current works often implement their attention mechanisms following GAT \cite{velivckovic2018graph}. However, GAT layers do not include edge-feature information and have been shown to only compute static attention coefficients. To address this problem, we design our attention score inspiring from GATv2 \cite{brody2021attentive} and make it to be relation-aware by using the pre-defined composition output as follow:
\begin{equation}
\label{eqn:attention}
    \alpha_{pqr}^k = \frac{\exp\left(\theta^k\left(\mathbf{h}_p^k, \mathbf{c}_{qr}^k \right)\right)}{\sum_{(q*, r*) \in \mathcal{N}(p)} \exp\left(\theta^k\left(\mathbf{h}_p^k, \mathbf{c}^k_{q*r*}\right)\right)}
\end{equation}
\begin{equation}
\theta^k\left(\mathbf{x}, \mathbf{y}\right) = \mathbf{a}^T LeakyReLU\left(\mathbf{W}^k_{att}\left[\mathbf{x} || \mathbf{y}\right]\right)
\end{equation}
where $\mathbf{W}^k_{att} \in \mathbb{R}^{d \times 2d}$ is a layer-wise attention weight matrix, $\mathbf{a} \in \mathbb{R}^d$ is an attention weight vector, and $||$ denotes concatenation. As the composition output vector $\mathbf{c}_{qr}^k$ contains the information of not only neighbor entity $q$ but also neighbor relation $r$, our attentive score $\alpha_{pqr}^k$ can capture the importance of the message coming from node $q$ to node $p$ conditioned on relation $r$ connecting them. 

\sititle{Final embdding} To achieve final embedding $\mathbf{h}_p$ for entity $p$ and $\mathbf{h}_r$ for relation $r$, we concatenate their embeddings at all layers and then use a linear transformation to project them to their final embedding space:
\begin{equation}
    \mathbf{h}_p = \mathbf{W}_{ge} [\mathbf{h}_p^0 || \mathbf{h}_p^1 || ... || \mathbf{h}_p^K]
    \label{eqn:all_ent}
\end{equation}
\begin{equation}
    \mathbf{h}_r = \mathbf{W}_{gr} [\mathbf{h}_r^0 || \mathbf{h}_r^1 || ... || \mathbf{h}_r^K]
    \label{eqn:all_rel}
\end{equation}
where $\mathbf{W}_{ge} \in \mathbb{R}^{d \times (K+1)d}$ and  $\mathbf{W}_{gr} \in \mathbb{R}^{d \times (K+1)d}$ is two linear transformation matrices.

\sititle{Loss function} We use the cosine distance metric to measure the distance between entities across knowledge graphs:
\begin{equation}
\label{eqn:dis}
    d_c\left(p, p' \right) = 1 - cos\left(\mathbf{h}_{p}, \mathbf{h}_{p'}\right)
\end{equation}
Our objective is to minimize the distance between aligned entity pairs while maximizing the distance between negative entity pairs, using a margin-based scoring function:
\begin{equation}
    \mathcal{L}_{gm} = \sum_{(p, p') \in \mathbb{L}}\sum_{(\bar p, \bar p') \in \mathbb{\bar L}} [\gamma_g + d_c(p, p') - d_c(\bar p, \bar p')]_+
    \label{eqn:g_m_loss}
\end{equation}
where $\gamma_g > 0$ is a margin hyper-parameter and $\mathbb{\bar L}$ is the set of negative entity pairs, which is constructed by replacing one end of each positively aligned pair by its close neighbours according to \autoref{eqn:dis} \cite{wu2019relation}.

Note that our \textit{relation-aware mechanisms} can only work if we can reconcile the relation embeddings as pointed out in \cite{sun2020knowledge}. Thanks to the aligned relation embeddings at the \textit{transitivity-based channel} we now can do this by adding the following loss term to transfer the relation alignment results from the \textit{transitivity-based channel} to the \textit{proximity-based channel}:
\begin{equation}
    \mathcal{L}_{gr} = \sum_{r \in \mathcal{R}} \sum_{r' \in \mathcal{R}}|d_{ct}(r, r') - d_{cg}(r, r')|
    \label{eqn:g_r_loss}
\end{equation}
where $d_{ct}(r, r')$ and $d_{cg}(r, r')$ are the cosine distances between $r$ and $r'$ w.r.t translation-based channel and \textit{proximity-based channel} relation embeddings respectively. 

We finally combine \autoref{eqn:g_m_loss} and \autoref{eqn:g_r_loss} to get the final loss function of the \textit{proximity-based channel}:
\begin{equation}
    \mathcal{L}_g = \mathcal{L}_{gm} + \beta_g \mathcal{L}_{gr}
    \label{eqn:loss2}
\end{equation}
where $\beta_g \in \mathbb{R}$ is a hyper-parameter weighting the importance of $\mathcal{L}_{gr}$.

\section{The complete alignment process}
\label{sec:process}

\subsection{Alignment instantiation}
\label{sec:alignment}
We use the cosine similarity matrix to compute the similarity between any two entities across KGs:
\begin{equation}
    Sim(p, p') = cos(\mathbf{h}_p, \mathbf{h}_{p'})
    \label{eqn:simi}
\end{equation}

We then use this function to compute the alignment matrix $\mathbf{S}$ for each channel where $\mathbf{S}(p, q) = Sim(p, q)$ is proportional to the probability that $p$ is aligned to $q$. Suppose $\mathbf{S}_t$ and $\mathbf{S}_g$ is the alignment matrix of the translation-based channel and the \textit{proximity-based channel} respectively. We then combine them to form the final alignment matrix as follow:
\begin{equation}
    \mathbf{S} = \beta_{tg} \mathbf{S}_t + \left(1 - \beta_{tg}\right) \mathbf{S}_g
    \label{eqn:alignment}
\end{equation}
here $\beta_{tg}$ is a balancing hyper-parameter.

We then use the greedy match algorithm \cite{greedymatch} to infer the matching entities from the alignment matrix $\mathbf{S}$. For a fair comparison, we also apply this algorithm to all the baselines.

\subsection{Missing triples recovery} 
\label{sec:recovery}

In a typical KGC problem, we are often given a set of incomplete triples in which two out of three elements are available ($\langle h, r, ? \rangle$ or $\langle h, ?, t \rangle$, $\langle ?, r, t \rangle$). The task is to predict the missing elements in these triples, which means we know beforehand the position of missing triples. The model thus only needs to predict the missing relation with the searching space size equal to the number of relation $|\mathcal{R}|$. 
However, our model aims to tackle a more challenging problem in which we do not know that information beforehand. As a result, the model has to find all the missing elements of all possible missing triples with the much larger searching space size of $|\mathcal{V}|^2 \times |\mathcal{R}|$.

To overcome this challenge, we add a two-step module into our model. Firstly, a neural network is built to find all pairs of entities that might have missing relations. We use a 2-layer MLP followed by a sigmoid function to return the probability of how likely some relations between any two entities were missed. For each entity pair $\langle p, q \rangle$, we compute the mentioned probability $p_e(p, q)$ for it as follow:
\begin{equation}
    p_e(p, q) = Sigmoid\left(MLP\left( \mathbf{h}_p || \mathbf{h}_q \right)\right)
    \label{eqn:pro_edge}
\end{equation}

After having $p_e(p, q)$, we compare this value with the average $\nu$ of all the probabilities of pairs which already have relations connecting them. If $p_e(p, q) \geq \nu$, this entity pair will be passed through the second step to predict the missing relations between them. In the second step, since we already have pairs of entities predicted to have missed relations, we only need to apply the tradition KGC methods as mentioned before. Suppose $\langle p, q \rangle$ have already predicted to have missing relations, then for each relation $r$, we compute the dissimilarity value $d_t(p + r, q)$. This value will be used to decide whether $\langle p, r, q \rangle$ should be filled to the graph. If the dissimilarity value is not larger than the average of this measure of all the positive triples $\eta$, which means $d(p + r, q) \leq \eta$, then $\langle p, r, q \rangle$ will be allowed to be filled to the graph. 

Note that we do not use the embeddings of the \textit{proximity-based channel} in this module because this module employs GNN with fixed initialized word embeddings for entities, which makes this module fail to capture positional information. As a consequence, this module would not perform well when tackling the knowledge completion task. Whereas on the other hand, the transitivity based channel is a shallow architecture with trainable entities representations with higher degree of freedom which allows the module to better capture positional information. Thus, we only make use of the first channel representation for filling missing triples to the graphs. As a result, We update the loss function of the \textit{Transitivity-based channel} \autoref{eqn:trans_loss} as follow:
\begin{equation}
    \mathcal{L}_t = \mathcal{L}_{ts} + \beta_{tc}  \mathcal{L}_{tc} + \beta_{tm} \mathcal{L}_{tm}
    \label{eqn:final_tran_loss}
\end{equation}
where $\beta_{tm}, \beta_{tc}$ are two real hyper-parameters weighting the importance of $\mathcal{L}_{tm}$ and $\mathcal{L}_{tc}$, respectively; $\mathcal{L}_{tc}$ is a loss function allowing the first \textit{Missing triples recovery} module to be trained to produce reasonable $p_e(.)$:
\begin{equation}
    \mathcal{L}_{tc} = \sum_{(p, q) \in \mathcal{P}} \left( - log\left(p_e\left(p, q\right)\right) + \sum_{q' \propto P_n\left(p\right)} log\left(p_e\left(p, q'\right)\right)  \right) 
\end{equation}
where $\mathcal{P}$ is the set of entity pairs that already have relations connecting them, and $P_n(p)$ is a set of negative examples (i.e., set of $p$'s non-neighbor entities). In our implementation, we first optimize $ (\mathcal{L}_{ts} + \beta_{tc} \mathcal{L}_{tc})$ and then optimize $\beta_{tm}\mathcal{L}_{tm}$ latter. 

The \textit{proximity-based channel} also indirectly contributes to the triples recovering process. Because it utilizes name and neighborhood structure information to enhance the module's matching quality. In turn,  this matching information can be used to better recover missing relations between entities. 

Suppose the model successfully recognized $p$ and $p'$, $q$ and $q'$, and $r$ and $r'$ are aligned together, $\langle p, r, q \rangle$ already in the source KG. Then even if $\langle p', r', q' \rangle$ is a missing triple in the second graph, our model can easily recovers it.

\sstitle{False links correction} During the missing links recovery process, the model might add wrong relations to the KGs when recovering missing triples. Consequently, this action could exert undesirable impacts on the model's performance. Thus, we propose a mechanism allowing our model to correct itself from wrong decisions, preventing it from accumulating noises to the KGs. The intuition is to keep checking the dissimilarity value $d_t(.)$ and missing edge probability $p_e(.)$ of the filled triples. If an added triple $\langle h, r, t \rangle$ fails to satisfy the recovery condition (i.e., $d(h + r, t) > \eta $ or $p(h, t) < \nu$) at any iteration, it will be removed from the filled triple set. Although this mechanism does not ensure all the wrong decisions would be detected and fixed, it allows the model to be more adaptive. 

\sstitle{Scaling to large input} To allow our proposed algorithm to scale well to large KGs, we introduce a relaxed version of \textit{Missing triple recovery}. Firstly, at each iteration, we uniformly sample a small subset $\mathcal{V}'$ of entities to perform the triple recovery step. Because the complexity of this step is quadratically proportional to the number of entities, this subsampling strategy will significantly reduce the running time of the model. Another benefit is that this mechanism will prevent the model from adding too many triples, most of which can be false as they are inferred from the poor, learned representations of KGs at the early stage of the training process. Consequently, this action may potentially accumulate too much noise to the graphs. Secondly, added triples could be huge as the running process iterates. As a consequence, the \textit{false links correction} step could cost a significant amount of time because the model has to check every single triple repeatedly. Thus, we allow the model to ignore triples that already pass more than $e$ consecutive checking steps, letting them stay permanently in the triple set.

\subsection{Link-augmented training process}

The training process for the whole model is depicted in \autoref{alg:training}. We preprocess the input KGs in line 3, then initialize the learnable parameters in the two channels in line 4. For the Transitivity-based channel, we initialize entity embeddings and relation embeddings by Xavier initialization. On the other hand, for Proximity-based channel, we employ the node features as pre-trained English word vectors trained by fastText \cite{mikolov2018advances}. This initialization is also applied to all the baselines that use entity name embedding. For each training step, we update the entity embeddings and relation embeddings in the two channels by minimizing the loss functions (see \autoref{eqn:loss1}, \autoref{eqn:loss2}) using Adam optimizer (line 6-11). Then, we periodically update the current alignment matrix $\mathbf{S}$ and nominate top $c$ matched entity pairs based on their similarity values. The candidates nominated at least $n$ times are considered as \textit{high-confidence anchor links} and updated to alignment seed set $\mathbb{L}$. Finally, the optimized matrix $\mathbf{S}$ is used to retrieve the aligned pairs. At each 10 epoch, we compute the similarity matrix using Eq.~\ref{eqn:alignment} and update the alignment seeds as mentioned earlier. We also evaluate our model's performance during training. The process will stop if the development MRR does not increase in two consecutive evaluation steps.

\begin{algorithm}
\scriptsize
\caption{Training scheme}
\label{alg:training}
\begin{algorithmic}[1]
\State \textbf{Input}: source and target input KGs: $\mathcal{KG}_s$ and $\mathcal{KG}_t$, entity alignment seeds $\mathbb{L}$.
\State \textbf{Output}: optimized alignment matrix $\mathbf{S}$
\State Add inverse and self loop edges to reach KG
\State Initialize the embeddings 
\For{epoch $e$ in $\{1,2,...,N\}$}
\State Update Translation based embedding to minimize \autoref{eqn:final_tran_loss}
\For{each GNN layer k}
\State Compute layer-wise entity emb. $\mathbf{h}^k_e$ using \autoref{eqn:entity_update}
\State Compute layer-wise relation emb. $\mathbf{h}^k_r$ using \autoref{eqn:relation_update}
\EndFor
\State Update final GNN embeddings for entities and relations \autoref{eqn:all_ent}, \autoref{eqn:all_rel} 
\State Update model parameter to minimize \autoref{eqn:loss2}
\If{ $e \equiv 1 \pmod{valid\_epoch}$}
\State Update the alignment matrix $\mathbf{S}$ using \autoref{eqn:alignment}
\State Update the alignment seeds $\mathbb{L}$ using $\mathbf{S}$ 
\State Recover some missing triples following \ref{sec:recovery}
\EndIf
\EndFor
\State \Return{$\mathbf{S}$}
\end{algorithmic}
\end{algorithm}

\subsection{Complexity Analysis} 
\label{sec:com}

To analyze the time complexity of our model, we will focus on different parts of our models. In the \textit{Transitivity-based channel}, the translation constraint loss costs $\mathcal{O}(|\mathcal{E}|)$, while the mapping loss costs $\mathcal{O}(|\mathcal{V}|)$. The triples recovery process takes $\mathcal{O}(|\mathcal{V}|^2 \times |\mathcal{R}|)$.

On the other hand, in the \textit{Proximity-based channel}, the relation-aware attention and relation-aware message passing costs $\mathcal{O}(\mathcal{E})$. Beside, the entities mapping loss and relation mapping loss cost $\mathcal{O}(|\mathcal{V}|)$ and $\mathcal{O}(|\mathcal{R}|^2)$ respectively. In the alignment computation step, the greedy match and high confidence sampler cost $\mathcal{O}(|\mathcal{V}|^3)$. 

In sum, the total time complexity is $\mathcal{O}(|\mathcal{E}| + |\mathcal{V}|^2 \times |\mathcal{R}| + |\mathcal{V}|^3)$

\section{Evaluation}
\label{sec:exp}

In this section, we report the experimental results of our techniques against a sizeable collection of 8 baselines and 8 real-world datasets, covering different aspects such as end-to-end comparison, ablation study, data sparsity, labelling effort, and qualitative evidence. 

\subsection{Experimental Setup}

\sstitle{Dataset} We use 4 popular benchmarking datasets from \cite{sun2020benchmarking}, including four datasets of two cross-lingual versions: English-French and English-German which were crawled from DBpedia (2016-2020) \cite{lehmann2015dbpedia}. The datasets have a total of 240K entities, 2438 relations, and 964543 relational triples. Each dataset consists of two versions; (v1) is the sparse version while (v2) is the dense one. Version V1 was directly obtained using the IDS algorithm \cite{suchanek2008yago}, while version V2 was created by filtering out low-degree entities and thus being denser than V1. The details of the datasets using in our experiments are shown in \autoref{tab:gmdataset}.

\begin{table}[!h]
\centering
\caption{Dataset statistics for KG alignment}
\label{tab:gmdataset}
\resizebox{\columnwidth}{!}{
\begin{tabular}{lr|ccc|ccc}
\toprule
\multirow{2}{*}{\bf Datasets} & \multirow{2}{*}{\bf KGs} & \multicolumn{3}{c|}{\bf V1} & \multicolumn{3}{c}{\bf V2} \\
\cline{3-8} 
& &  \#Ent. & \#Rel.  & \#Rel tr.  &  \#Ent. &  \#Rel. & \#Rel tr.  \\
\midrule
\multirow{2}{*}{EN-FR-15K} & EN & 15,000 & 267 & 47,334 & 15,000 & 193 & 96,318 \\
& FR & 15,000 & 210 & 40,864 & 15,000 & 166 & 80,112 \\
\midrule
\multirow{2}{*}{EN-DE-15K} & EN & 15,000 & 215 & 47,676 & 15,000 & 169 & 84,867 \\
& DE & 15,000 & 131 & 50,419 & 15,000 & 96 & 92,632 \\
\midrule
\multirow{2}{*}{D-W-15K} & DB & 15,000 & 248 & 38,265 & 15,000 & 167 & 73,983 \\
& WD & 15,000 & 169 & 42,746 & 15,000 & 121 & 83,365 \\
\midrule
\multirow{2}{*}{D-Y-15K} & DB & 15,000 & 165 & 30,291 & 15,000 & 72 & 68,063 \\
& YG & 15,000 & 28 & 26,638 & 15,000 & 21 & 60,970 \\
\hline
\bottomrule
\end{tabular}}
\end{table}


\sstitle{Baselines} We compare the performance of our techniques with 8 SOTA alignment techniques as follows:

\begin{enumerate}
\item MTransE \cite{bordes2013translating} is a transitivity-based entity alignment model which employs the translation constraint to learn the representation for each entity and relation in the input KGs using a shallow embedding model. 
\item JAPE \cite{sun2017cross} is a shallow embedding-based technique that generates \textit{structure embedding} (using a TransE model) and \textit{attribute embedding} (using a skip-gram model). The learnt embeddings are then used simultaneously to align the entities. 

\item KDcoE \cite{chen2018co} leverages co-training of a KG embedding and a literal description embedding to enhance the semi-supervised learning of multilingual KG embeddings. The KG embedding model jointly trains a translation model with a linear-transformation-based alignment, while the description embedding model employs an attentive GRU encoder (AGRU) to characterize multilingual entity descriptions. 

\item GCNAlign \cite{wang2018cross} is a deep embedding-based technique that employs two GCNs to produce structural and attribute embeddings, which capture the relational and attribute information of the entities. Both embeddings are then used to discover corresponding entities via an alignment seed set~\cite{wang2018cross}. 

\item BootEA \cite{sun2018bootstrapping} is a shallow embedding-based technique that generates entity and relation embeddings using translation constraint like MTransE. The model then reconciles the learnt embeddings using alignment seeds, followed by an alignment editing step to reduce the accumulated errors~\cite{sun2018bootstrapping}. 

\item MultiKE \cite{zhang2019multi} is an embedding-based framework that unifies multiple views of entities to learn embeddings for entity alignment. The model implements three representative views corresponding to name, relation and attribute features, respectively; then entity
alignment are retrieved by combining the learnt embeddings in three different strategies. 

\item RDGCN \cite{wu2019relation} is a deep embedding-based technique that employs a two-layers GCN with highway gates to capture the attentive interaction between each KG and its dual relation. The embeddings are then compared directly to obtain the alignment result~\cite{wu2019relation}.

\item Alinet \cite{sun2020knowledge} is a deep embedding-based technique that leverages entities' distant neighbours to preprocess the input KGs and expand their neighbour overlapping. The technique then forwards the processed inputs through an attention-based GNN which controls the aggregation of both direct and distant neighborhood information using a gating mechanism.


\end{enumerate}

\sstitle{Evaluation metrics}
In our experiments, we consider alignment direction from left to right following the setting defined in \cite{sun2020benchmarking}. We use Hit@$m$ ($m = 1, 10$) to measure the prediction ability of the techniques based on how accurately true positive alignments are observed in the top-k candidates \cite{userIdentityBenchmark}. Suppose $\mathbb{L}(p) \in \mathcal{V}_t$ is the counterpart of entity $p \in \mathcal{V}_s$:
\begin{equation}
\label{eqn:hit}
Hit@m = \frac{\sum_{p \in \mathcal{V}_s} \mathds{1}_{\text{$\mathbf{S}(p,\mathbb{L}(p))$ $\in$ top-$m$ $\mathbf{S}(p)$}}}{\#\{\text{True anchor links}\}}
\end{equation}
where $\mathds{1}_{hypo} = 1$ if $hypo$ is true, and $\mathds{1}_{hypo} = 0$ if $hypo$ is false. 


We also use Mean Reciprocal Rank (MRR) and Mean Rank (MR) under pair-wise setting to measure how high the true anchored links are ranked in the list of alignment candidates: 
\begin{align}
MRR = mean \left(\frac{1}{tr} \right) \\
    MR = mean(tr)
\end{align}
where $tr$ is the rank of the matched score of the true target. MR and MRR metrics measure how the models perform at soft alignment scenarios where most of the true target entities have high similarity with the source entities. 

\sstitle{Hyper-parameter setting}
Our multi-channel model requires in total of 16 hyperparameters, which is reported in \autoref{tab:hyper} for reproducibility. 

\begin{table}[!h]
    \centering
    \footnotesize
    \caption{Hyper-parameter setting}
    \label{tab:hyper}
    \resizebox{0.8\columnwidth}{!}{
    \begin{tabular}{l | r}
    \toprule
    Hyper-parameter name & Hyper-parameter value \\
    \midrule
    Embedding dim & 100 \\
    Learning rate & 0.1 \\
    Mapping weight $\beta_t$ & 50 \\
    Batch size & 1000 \\
    Optimizer & Adam \\
    \#negative samples & 5 \\
    Margin $\gamma_t$ & 1 \\
    Embedding dim & 300 \\
    \#GCN layers $K$ & 2 \\
    Hidden dim & 300 \\
    \#negative samples & 50 \\
    Learning rate & 0.0005 \\
    Attention Leaky ReLU $w$ & 0.05 \\
    Margin $\gamma_g$ & 1 \\
    Similarity balancing weight $\beta_{tg}$ & 0.4 \\
    Eval step & 10 \\
    \bottomrule
    \end{tabular}
    }
\end{table}

\sstitle{Computational environment} 
The experiments were averaged over 50 runs for each dataset to mitigate the effect of randomness. We used an AMD Ryzen ThreadRipper 3.8 GHz system with 64 GB RAM and four GTX Titan X graphic cards. We implemented our model in Python with Pytorch library.

\subsection{End-to-end comparison}

\begin{table*}[!h]
\centering
\caption{End-to-end KG alignment performance (bold: winner, underline: first runner-up)}
\label{tab:EA_end-to-end}
\resizebox{1.6\columnwidth}{!}{
\begin{tabular}{@{} l | l | l | l l l l l l l l l@{}}
\toprule
\textbf{Dataset} & \textbf{Ver.} & \textbf{Metric} & \textbf{MTransE} & \textbf{GCN-A} & \textbf{BootEA} & \textbf{RDGCN} & \textbf{Alinet} & \textbf{JAPE} & \textbf{KDcoE} & \textbf{MultiKE} & \textbf{IKAMI} \\
\midrule
\multirow{8}{*}{\textbf{EN-DE}} & \multirow{4}{*}{\textbf{V1}} & Hit@1 & .307 & .481 & .675 & \underline{.830} & .609 & .288 & .529 & .756 & \bf .949 \\
& & Hit@10 & .610 & .753 & .865 & \underline{.915} & .829 & .607 & .679 & .828 & \textbf{.991} \\
& & MRR & .407 & .571 & .740 & \underline{.859} & .681 & .394 & .580 & .782 & \bf .952 \\
& & MR & 223.9 & 352.3 & 125.7 & \underline{67.1} & 216.7 & 140.6 & 124.8 & 91.5 & \textbf{8.4} \\
\cline{2-12} 
& \multirow{4}{*}{\textbf{V2}} & Hit@1 & .193 & .534 & \underline{.833} & \underline{.833} & .816 & .167 & .649 & .755 & \bf .964 \\
& & Hit@10 & .431 & .780 & \underline{.936} & \underline{.936} & .931 & .415 & .835 & .835 &\bf .992 \\
& & MRR & .274 & .618 & .869 & \underline{.860} & .857 & .250 & .715 & .784 & \bf .975 \\
& & MR & 193.5 & 108.0 &  16.2 & 74.8 & 71.1 & 139.9 & \underline{16.0} & 45.2 & \textbf{3.0} \\
\midrule
\multirow{8}{*}{\textbf{EN-FR}} & \multirow{4}{*}{\textbf{V1}} & Hit@1 & .247 & .338 & .507 & \underline{.755} & .387 & .263 & .581 & .749 & \bf .907 \\ 
& & Hit@10 & .563 & .680 & .794 & \underline{.880} & .829 & .595 & .721 & .843 & \bf .992 \\
& & MRR & .351 & .451 & .603 & \underline{.800} & .487 & .372 & .628 & .782 & \bf .935 \\
& & MR & 251.9 & 562.2 & 227.7 & 156.1 & 483.2 & 175.6 & 197.0 & \underline{97.8} & \textbf{7.2} \\
\cline{2-12} 
& \multirow{4}{*}{\textbf{V2}} & Hit@1 & .240 & .414 & .660 & \underline{.847} & .580 & .292 & .730 & .864 & \bf .978 \\ 
& & Hit@10 & .240 & .796 & .906 & \underline{.934} & .877 & .624 & .869 & .924 & \textbf{.998} \\
& & MRR & .336 & .542 & .745 & \underline{.880} & .689 & .402 & .778 & .885 & \bf .986 \\ 
& & MR & 206.0 & 131.3 & 25.7 & 61.7 & 94.0 & 89.1 & 27.3 & \underline{12.1} & \textbf{1.2}  \\
\midrule
\multirow{8}{*}{\textbf{D-W}} & \multirow{4}{*}{\textbf{V1}} & Hit@1 & .259 & .364 & \underline{.572} & .515 & .470 & .250 & .247 & .411 & \bf .724 \\ 
& & Hit@10 & .541 & .648 & \underline{.793} & .717 & .703 & .541 & .473 & .583 & \textbf{.911} \\
& & MRR & .354 & .461 & \underline{.649} & .584 & .552 & .348 & .325 & .468 & \bf .793 \\ 
& & MR & 331.1 & 765.3 & 286.3 & 508.5 & 575.7 & \underline{243.7} & 730.2 & 275.4 & \textbf{25.3} \\
\cline{2-12} 
& \multirow{4}{*}{\textbf{V2}} & Hit@1 & .271 & .506 & \underline{.821} & .623 & .741 & .262 & .405 & .495 & \bf .857 \\ 
& & Hit@10 & .584 & .818 & \underline{.950} & .805 & .925 & .581 & .720 & .724 & \textbf{.984} \\
& & MRR & .376 & .612 & \underline{.867} & .684 & .807 & .368 & .515 & .569 & \bf .900 \\ 
& & MR & 146.0 & 146.0 & \underline{18.4} & 229.3 & 72.1 & 99.0 & 91.7 & 38.6 & \textbf{3.0} \\
\midrule
\multirow{8}{*}{\textbf{D-Y}} & \multirow{4}{*}{\textbf{V1}} & Hit@1 & .463 & .465 & .739 & \underline{.931} & .569  & .469 & .661 & .903 & \bf .967 \\ 
& & Hit@10 & .733 & .661 & .871 & \underline{.974} & .726 & .747 & .797 & .950 & \textbf{.990} \\
& & MRR & .559 & .536 & .788 & \underline{.949} & .630 & .567 & .710 & .920 & \bf .976 \\ 
& & MR & 245.6 & 1113.7 & 365.1 & \underline{17.8} & 532.6  & 211.2 & 133.3 & 19.5 & \textbf{3.1} \\
\cline{2-12}
& \multirow{4}{*}{\textbf{V2}} & Hit@1 & .443 & .875 & .958 & \underline{.936} & .951 & .945 & .895 & .856 & \bf .987 \\ 
& & Hit@10 & .707 & .963 & \underline{.990} & .973 & .989 & .626 & .984 & .927 & \textbf{.998} \\
& & MRR & .533 & .907 & \underline{.969} & .950 & .965 & .440 & .932 & .881 & \bf .992 \\ 
& & MR & 85.2 & 47.1 & 4.8 & 13.8 & 5.6 & 82.5 & \underline{2.1} & 10.0 & \textbf{1.1 }\\
\bottomrule
\end{tabular}
}
\end{table*}

\begin{table}[!h]
\begin{minipage}[c]{\linewidth}
\caption{Ablation study}
\label{tab:ablation_EA}
\vspace{0em}
\centering
\resizebox{1.0\columnwidth}{!}{
\begin{tabular}{l|ccc|ccc}
\toprule
\multirow{2}{*}{Var} & \multicolumn{3}{c|}{\bf D-W-V1} & \multicolumn{3}{c}{\bf D-W-V2} \\
\cline{2-7}
& Hit@1 & Hit@10  & MRR & Hit@1 & Hit@10 & MRR  \\
\midrule

\textit{Var1} & .685 & .863 & .750 & .784 & .942 & .841 \\
\textit{Var2} & .691 & .883 & .762 & .818 & .962 & .870 \\
\textit{Var3} & .716 & .903 & .783 & \underline{.828} & .960 & .873 \\
\textit{Var4} & \underline{.722} & \underline{.908} & \underline{.791} &  .822 & \underline{.970} & \underline{.876} \\
\textit{Var5} & .421 & .741 & .498 & .512 & .782 & .641 \\
\textit{Var6} & .468 & .752 & .515 & .556 & .799 & .663 \\
\textit{Var7} & .379 & .639 & .468 & .628 & .875 & .712 \\

\midrule
Full model & \bf .724 & \bf .911 & \bf .793 & \bf .832 & \bf .974 & \bf .883 \\
\bottomrule
\end{tabular}}

\vspace{1em}

\centering
\caption{Knowledge Graph Completion performance}
\label{tab:KGC_end-to-end}
\vspace{0em}
\resizebox{1.0\columnwidth}{!}{
\begin{tabular}{@{} l | l | l | l l l l l@{}}
\toprule
\textbf{Dataset} & \textbf{Ver.} & \textbf{Metric} & \textbf{DistMult} & \textbf{TransE} & \textbf{RotatE} & \textbf{CompGCN} & \textbf{IKAMI} \\
\midrule
\multirow{4}{*}{\textbf{EN-FR}} & \multirow{2}{*}{\textbf{V1}} & Hit@1 & .177 & .239 & .251 & \underline{.324} & \bf .485 \\ 
& & MRR & .311 & .365 & .381 & \underline{.421} & \bf .621 \\ 
\cline{2-8} 
& \multirow{2}{*}{\textbf{V2}} & Hit@1 & .193 & .195 & .205 & \underline{.314} & \textbf{.329} \\ 
& & MRR & .337 & .340 & .357 & \underline{.413} & \bf .469 \\ 
\midrule
\multirow{4}{*}{\textbf{EN-DE}} & \multirow{2}{*}{\textbf{V1}} & Hit@1 & .042 & .041 & .048 & \bf .234 & \underline{.148} \\
& & MRR & .089 & .102 & .120 & \bf .318 & \underline{.248} \\
\cline{2-8} 
& \multirow{2}{*}{\textbf{V2}} & Hit@1 & .122 & .125 & .124 & \underline{.187} & \bf .261 \\
& & MRR & .199 & .202 & .207 & \underline{.258} & \bf .369 \\
\bottomrule
\end{tabular}}

\end{minipage}
\end{table}

We report an end-to-end comparison of our alignment model against baseline methods on the real-world datasets in \autoref{tab:EA_end-to-end}. It can be seen that our model outperformed the others in all scenarios. Though using a multi-channel mechanism as \textit{GCN-Align} and \textit{RDGCN}, the gain of up to 10-20\% of Hit@1 and MRR demonstrated the efficiency of relation-aware integration and knowledge transfer mechanism proposed in our work \textit{IKAMI}, especially for denser version (v2) of the datasets. Also, we achieved much higher results than the transitivity-based model MTransE, which justified the superiority of our proximity GNN-based model. 

Among the baselines, RDGCN and BootEA, the two deep embedding-based techniques, were the runner-ups. Overall, they achieved up to 93.6\% of Hit@1 and 0.969 of MRR over all settings, except the noisy D-W dataset. AliNet and GCN-Align also gave promising results, which demonstrates the power of graph neural networks for entity alignment. On the other hand, the shallow embedding-based method MTransE achieved the lowest values for accuracy. 

\begin{figure}
    \centering
    \includegraphics[scale=0.33]{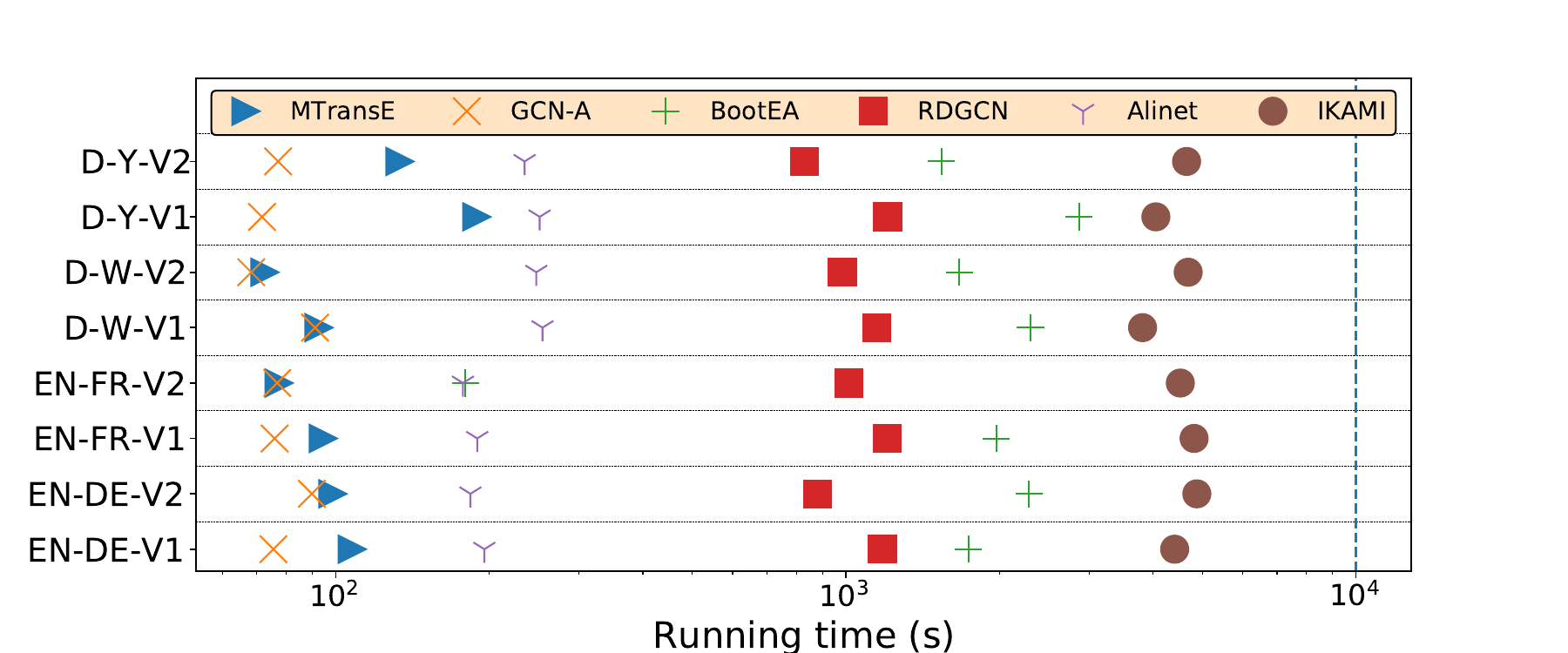}
    \caption{Running time (in log scale) on different datasets}
    \label{fig:time}
\end{figure}

When it comes to the scalability, \autoref{fig:time} depicts the running time of the techniques. \emph{GCN-Align} is the fastest because of full-batch setting of GCN. Our model requires the most running time, due to the accuracy-running time trade-off. Note that our framework allows the users reducing the number of iteration of completion and alignment improvement to reduce the time in sacrificing the alignment accuracy.

\subsection{Ablation Study}

We evaluate the design choices in our model by comparing the performance of the final model with several variants. The variants we consider are:
\begin{itemize}
    \item \emph{Var1:} removes the Transitivity-based channel and only keeps the Proximity-based channel
    \item \emph{Var2:} removes the Proximity-based channel and only keeps the Translation-based channel
    \item \emph{Var3:} replaces our Proximity-based channel architecture by an original GCN. \cite{kipf2016semi}
    \item \emph{Var4:} replaces our Proximity-based channel architecture by RGCN \cite{chen2019rgcn}.
    \item \emph{Var5:} does not contain the attention mechanism described in Sec. 3.3.
    \item \emph{Var6:} does not contain the updating alignment seed described in Sec. 3.4.
    \item \emph{Var7:} does not contain the swapping mechanism described in Sec. 3.1. 
\end{itemize}

We conduct this experiment in the D-W-V1 and D-W-V2 datasets. As from \autoref{tab:ablation_EA}, the full \emph{IKAMI} model outperformed the other variants in both datasets. In more detail, the full model achieved around 5\% and 35\% higher of Hit@1 and MRR comparing to the single-channel variants \textit{Var1} and \textit{Var2}, which shows advantages of fusing the two channels and confirms that the add of transitivity channel helps to complete the proximity GNN-based channel. Also, the gain of nearly 30\% of \textit{IKAMI} over the variant \textit{Var3} demonstrates the robustness of the innovations introduced in our GNN-based channel comparing to the original GCN. The performance witnessed a dramatic drop when we replaced our GNN architecture with R-GCN (\textit{var4}). Also, the lack of each \textit{attention mechanism} (\textit{var5}), \textit{alignment seed update} (\textit{var6}) and \textit{swapping mechanism} (\textit{var7}) caused a minor decrease in all the five metrics comparing to the full model. This proves the importance of these technical innovations to our model.

\subsection{Robustness to KGs incompleteness}

We evaluate the robustness of our method against the incompleteness by first investigating the capability of our embeddings to discover missing links (a.k.a KG completion \cite{wang2014knowledge}). To this end, from the original KGs pair, we randomly removed 20\% triples from the source graph. Then, we recovered the missing triples by selecting the tail entity $t$ that had the closet embeddings to the querying head entity and the relation pair $\langle h, r \rangle$ and vice-versa. We compared \textit{IKAMI} against four baseline KG completion techniques, namely \textit{DistMult} \cite{dettmers2018convolutional}, \textit{RotatE} \cite{nguyen2018novel}, \textit{TransE} \cite{bordes2013translating} and \textit{CompGCN} \cite{vashishth2019composition}. The result is shown in \autoref{tab:KGC_end-to-end}. It can be seen that \textit{IKAMI} were either the winner or the first runner-up, despite that our technique was not specialized for this task. Feature exchange between the proximity and transitivity channel can help to reconcile the KGs, which helps to reveal unseen relations from one graph based on similar patterns on the other. As we do not focus heavily on KG completion, interested readers can refer to other baselines~\cite{qiao2021context,che2020parame,zhang2020few}.

\begin{figure*}[!h]
    \centering
      \begin{subfigure}{0.7\linewidth}
        \centering
        \includegraphics[scale=0.15]{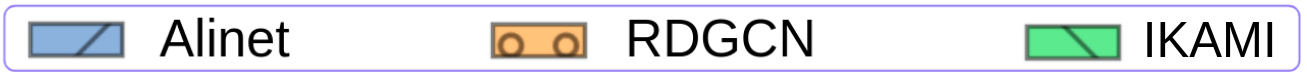}
        \label{fig:noise_legend}
        \end{subfigure}
      \begin{subfigure}{0.32\linewidth}
        \centering
        \includegraphics[scale=0.34]{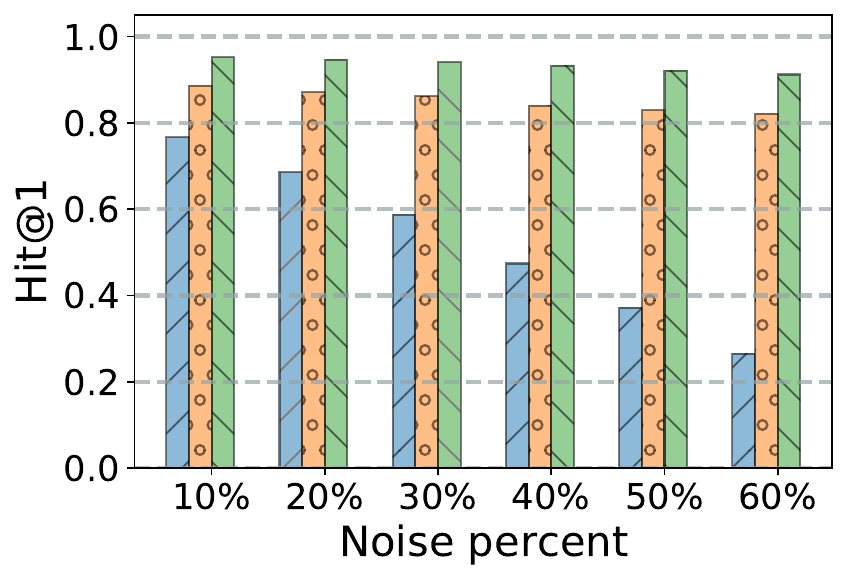}
        \caption{Hit@1}
        \label{fig:noise_hit1}
        \end{subfigure}
      \begin{subfigure}{0.32\linewidth}
        \centering
        \includegraphics[scale=0.34]{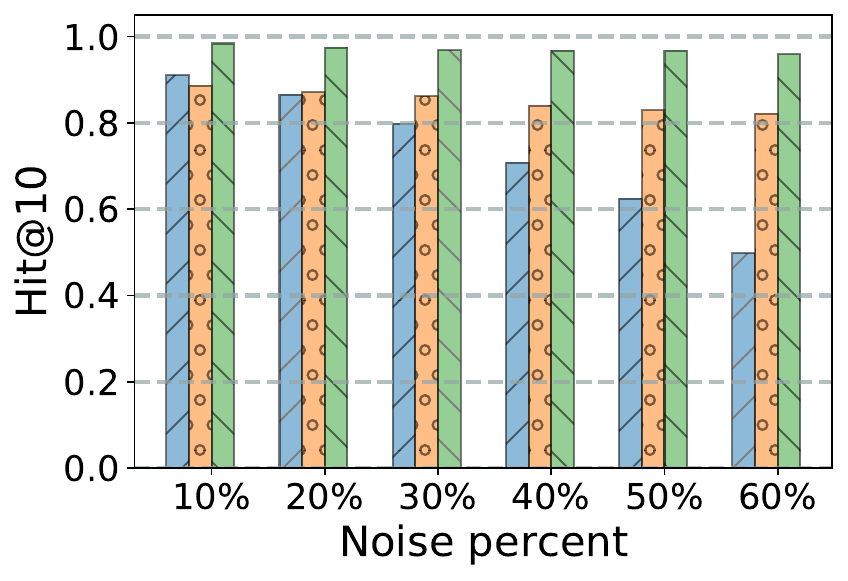}
        \caption{Hit@10}
        \label{fig:noise_hit10}
        \end{subfigure}
    \begin{subfigure}{0.32\linewidth}
        \centering
        \includegraphics[scale=0.34]{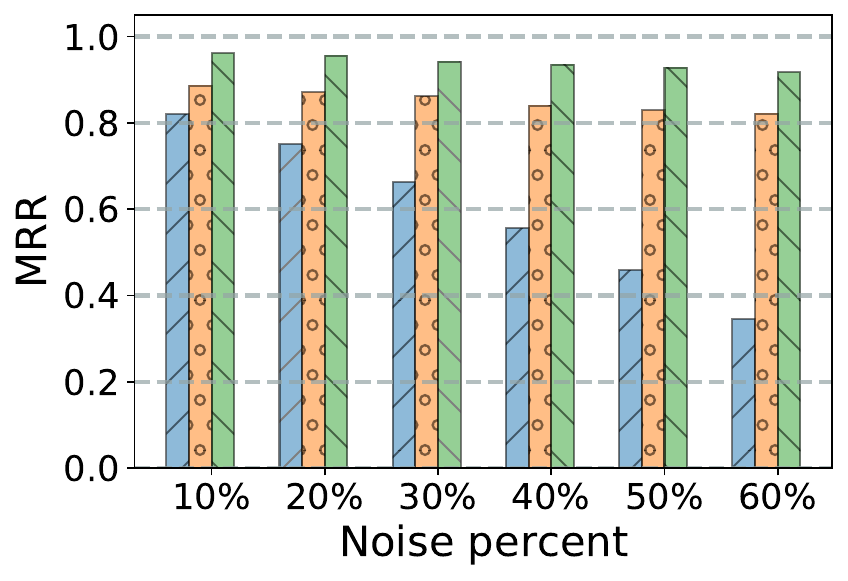}
        \caption{MRR}
        \label{fig:noise_mrr}
        \end{subfigure}
    \caption{Robustness of graph alignment models against noise on EN-DE-V2 test set}
    \label{fig:edge_noise}
 \end{figure*}

To fully investigate the robustness of the techniques against the KGs incompleteness, we conduct the second experiment where we choose the D-W-V2 as source KG and generate the target KG by removing the triples randomly to generate different levels of noise. The result of the experiment is shown in \autoref{fig:edge_noise}, where we only show the performance of IKAMI and the two best baselines RDGCN and AliNet. In general, all methods suffer performance drop when the noise level increases. Our model outperforms the baseline methods, with the Hit@1 goes from nearly $96\%$ to around $92\%$ when the edges removal ratio goes from $10\%$ to $60\%$, thanks to the efficient feature exchange mechanism. Our model keeps a margin of about $5\%$ in Hit@1 with the runner-up (RDGCN). The performance of Alinet drops more dramatically than the others, with less than 0.3 of Hit@1 and 0.5 of Hit@10 when the noise level goes up to 60\%.

\subsection{Saving of labelling effort}

\begin{figure*}[!h]
    \centering
      \begin{subfigure}{0.9\linewidth}
        \centering
        \includegraphics[scale=0.23]{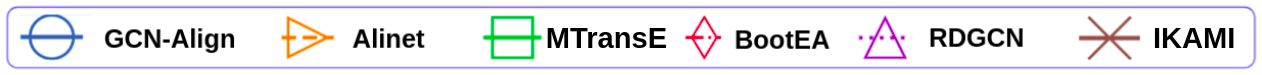}
        \label{fig:neg_size}
        \end{subfigure}
      \begin{subfigure}{0.32\linewidth}
        \centering
        \includegraphics[scale=0.35]{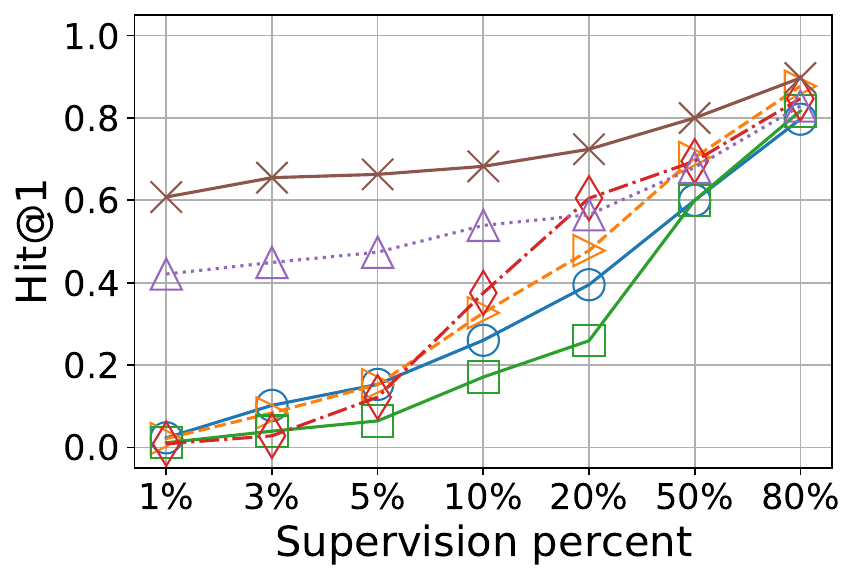}
        \caption{Hit@1}
        \label{fig:sub_hit1}
        \end{subfigure}
      \begin{subfigure}{0.32\linewidth}
        \centering
        \includegraphics[scale=0.35]{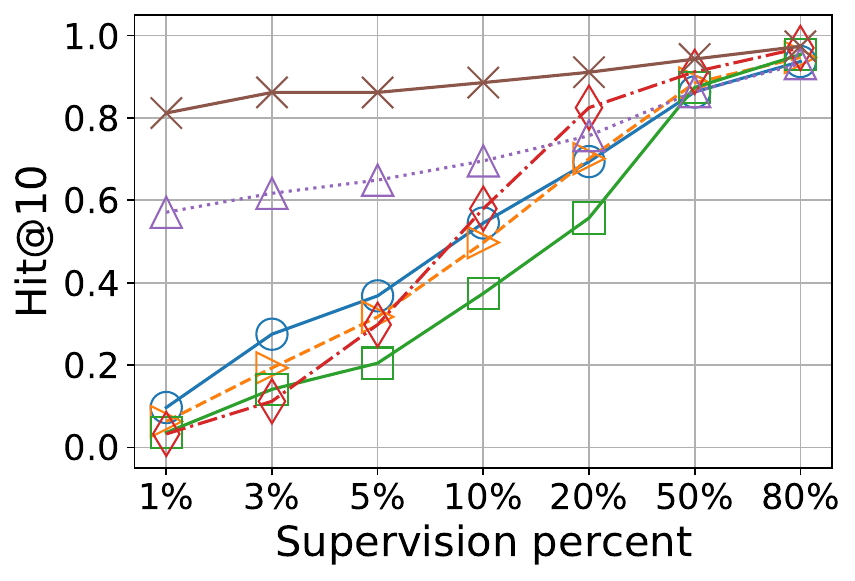}
        \caption{Hit@10}
        \label{fig:sub_hit10}
        \end{subfigure}
    \begin{subfigure}{0.32\linewidth}
        \centering
        \includegraphics[scale=0.35]{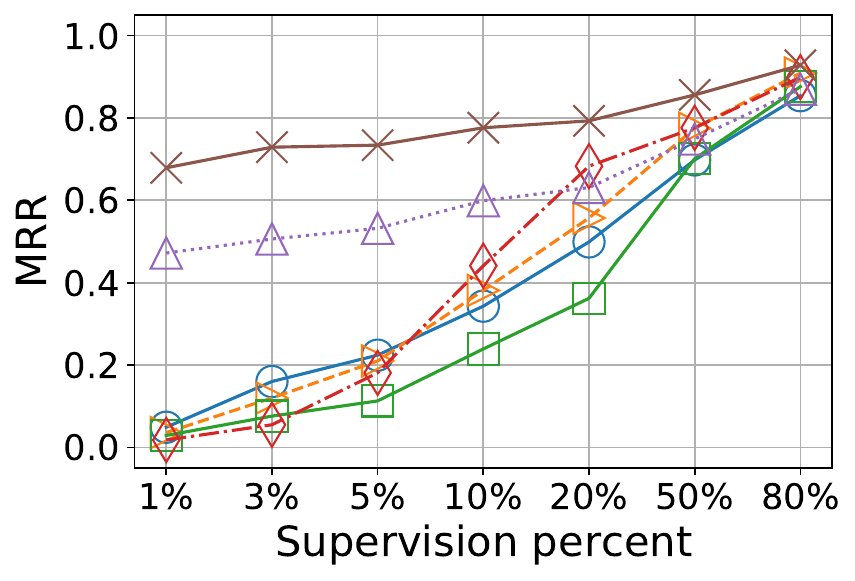}
        \caption{MRR}
        \label{fig:sub_mrr}
        \end{subfigure}
    \caption{Saving of labelling effort for entity alignment on D-W-V1 test set}
    \label{fig:supervision_percent}
 \end{figure*}

In this experiment, we evaluated the ability of saving pre-aligned entities of the techniques by examining their performance under different level of supervision data for the D-W dataset. It can be seen from \autoref{fig:supervision_percent} that our model \textit{IKAMI}  outperformed other baselines for every level of supervision, especially for the lower ones. We achieved a gain of around 20\% for the level of 1\% comparing to the second best baseline RDGCN. This result demonstrates the capability of the knowledge transform between the KGs using in \textit{IKAMI} in terms of saving labelling effort.

\subsection{Qualitative evidences}

\begin{figure}[!h]
    \centering
    \includegraphics[scale=0.45]{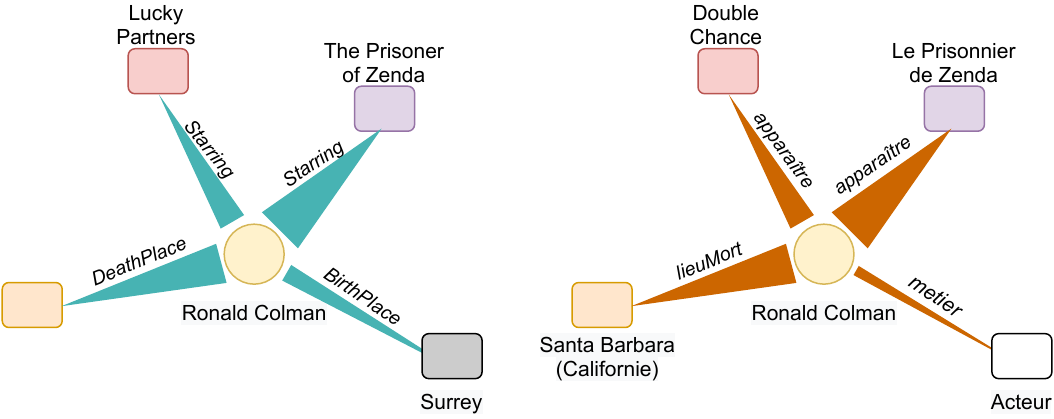}
    \caption{Attention visualisation (EN-FR-V1 dataset). The model pays less attention to noisy relations. }
    \label{fig:lp_attention}
\end{figure}

\begin{table}[!t]
\centering
\footnotesize
\caption{Correct aligned relations in EN$\leftrightarrow$FR KGs}
\label{tab:rel_align}
\resizebox{1.0\columnwidth}{!}{
\begin{tabular}{l}
\toprule
country $\leftrightarrow$ pays (country),
birthPlace $\leftrightarrow $ lieuNaissance (birth place), \\
deathPlace $\leftrightarrow$ lieuMort (dead place) 
starring $\leftrightarrow$ apparaître (starring) , \\
field $\leftrightarrow$ domaine (domain) ,
developer $\leftrightarrow$ développeurs (developer) \\
hometown $\leftrightarrow$ nationalité (nationality) \\
\bottomrule
\end{tabular}}
\end{table}

In this section, we qualitatively interpret our technique by two case studies. First, we visualized the attention coefficient of the relational triples of the entity Ronald Colman in~\autoref{fig:example} processed by \textit{IKAMI}. It is clear from \autoref{fig:lp_attention} that the coefficient for the triples appearing in both KGs outweighed that of the triples appearing in only one KG (e.g. \textit{BirthPlace} triple, \textit{Profession} triple). This depicts the capability of our attention mechanism in emphasizing the shared relational triples while mitigating the impact of the noisy ones. Second, list some representative relation alignment generated by the relation embedding from \textit{IKAMI} between EN and FR KGs. Our technique efficiently captured the underlying semantic of the relation type and aligned them quite accurately, without the need of machine translation. This also highlights the advantage of our relation representation learning and relation-aware propagation. 

\begin{figure}[!h]
    \centering
    \includegraphics[scale=0.4]{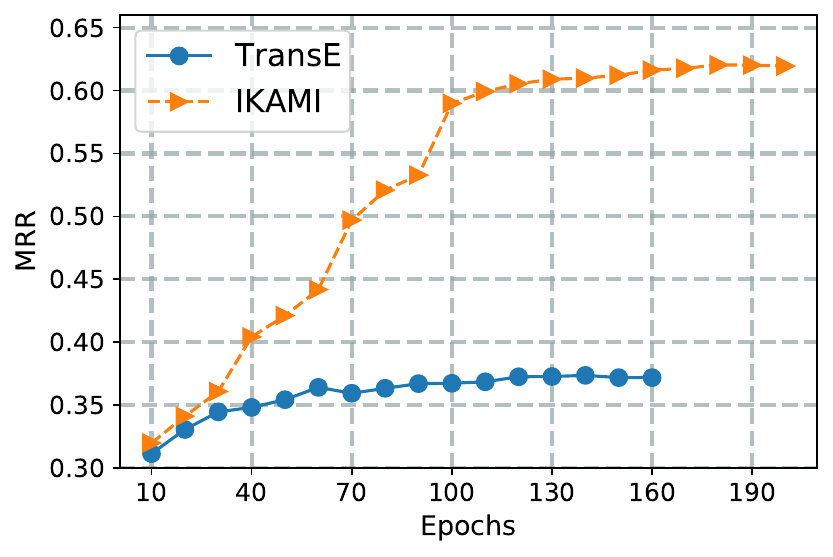}
    \caption{KGC performance comparison between TransE and IKAMI during training}
    \label{fig:lp_changing}
 \end{figure}

Second, we compare the KGC performance of \textit{IKAMI} with the single-channel transitivity-based technique TransE during the training process. It can be seen from \autoref{fig:lp_changing} that the fusion with proximity-based channel helps \textit{IKAMI} not only converged faster but also achieved superior final result against TransE.

\section{Conclusion}
\label{sec:con}

\subsubsection{Discussions}
We provide further key insights as follows: (1) IKAMI can achieve at least 70\% performance across all metrics with only 10\% of label information, where other baselines fail. (2) IKAMI can still maintain a 90\% performance across all metrics with 60\% of missing edges, where other baselines fail. (3) The ``seem-to-be-strict'' translation-based constraint \cite{bordes2013translating} surprisingly helps to strengthen the local signal and thus well-complete GNN-based model. (4) Unlike GNN-based existing works that often infer the relation embedding from entities, the relations should be assigned their own representation. Our relation-aware model in fact can learn underlying semantic and correctly align the relation between cross-lingual KGs without the need of machine translation. 

\subsubsection{Summary}
In this paper, we have presented a representation learning framework, IKAMI, for aligning incomplete knowledge graph from different domains. By exchanging multiple feature channels, including transitivity-based features and proximity-based features, between input knowledge graphs via representation learning, the alignment process is performed efficiently and overcomes the heterogeneity and incompleteness of KGs. Experiments show that our method improves various down-stream performances over SOTAs, including alignment, completion, sparsity, and labeling cost. In future work, we plan to incorporate external sources of information (e.g. transfer learning~\cite{song2021robust,tseng2021toward}) to further improve the alignment.


\ifCLASSOPTIONcompsoc
  \section*{Acknowledgments}
\else
  \section*{Acknowledgment}
\fi

Tong Van Vinh was funded by Vingroup Joint Stock Company and supported by the Domestic Master/ PhD Scholarship Programme of Vingroup Innovation Foundation (VINIF), Vingroup Big Data Institute (VINBIGDATA), code VINIF.2020.ThS.BK.07.


\begin{thebibliography}{10}
\providecommand{\url}[1]{#1}
\csname url@samestyle\endcsname
\providecommand{\newblock}{\relax}
\providecommand{\bibinfo}[2]{#2}
\providecommand{\BIBentrySTDinterwordspacing}{\spaceskip=0pt\relax}
\providecommand{\BIBentryALTinterwordstretchfactor}{4}
\providecommand{\BIBentryALTinterwordspacing}{\spaceskip=\fontdimen2\font plus
\BIBentryALTinterwordstretchfactor\fontdimen3\font minus
  \fontdimen4\font\relax}
\providecommand{\BIBforeignlanguage}[2]{{%
\expandafter\ifx\csname l@#1\endcsname\relax
\typeout{** WARNING: IEEEtran.bst: No hyphenation pattern has been}%
\typeout{** loaded for the language `#1'. Using the pattern for}%
\typeout{** the default language instead.}%
\else
\language=\csname l@#1\endcsname
\fi
#2}}
\providecommand{\BIBdecl}{\relax}
\BIBdecl

\bibitem{wang2017knowledge}
Q.~Wang, Z.~Mao, B.~Wang, and L.~Guo, ``Knowledge graph embedding: A survey of
  approaches and applications,'' \emph{TKDE}, vol.~29, no.~12, pp. 2724--2743,
  2017.

\bibitem{sun2020benchmarking}
Z.~Sun, Q.~Zhang, W.~Hu, C.~Wang, M.~Chen, F.~Akrami, and C.~Li, ``A
  benchmarking study of embedding-based entity alignment for knowledge
  graphs,'' \emph{Proc. VLDB Endow.}, vol.~13, no.~12, p. 2326–2340, 2020.

\bibitem{gracious2021neural}
T.~Gracious, S.~Gupta, A.~Kanthali, R.~M. Castro, and A.~Dukkipati, ``Neural
  latent space model for dynamic networks and temporal knowledge graphs,'' in
  \emph{AAAI}, vol.~35, no.~5, 2021, pp. 4054--4062.

\bibitem{trung2020adaptive}
H.~T. Trung, T.~Van~Vinh, N.~T. Tam, H.~Yin, M.~Weidlich, and N.~Q.~V. Hung,
  ``Adaptive network alignment with unsupervised and multi-order convolutional
  networks,'' in \emph{IEEE 36th International Conference on Data Engineering
  (ICDE)}, 2020, pp. 85--96.

\bibitem{phan2018pair}
M.~C. Phan, A.~Sun, Y.~Tay, J.~Han, and C.~Li, ``Pair-linking for collective
  entity disambiguation: Two could be better than all,'' \emph{TKDE}, vol.~31,
  no.~7, pp. 1383--1396, 2018.

\bibitem{wan2021gaussianpath}
G.~Wan and B.~Du, ``Gaussianpath: A bayesian multi-hop reasoning framework for
  knowledge graph reasoning,'' in \emph{AAAI}, vol.~35, no.~5, 2021, pp.
  4393--4401.

\bibitem{yan2021dynamic}
Y.~Yan, L.~Liu, Y.~Ban, B.~Jing, and H.~Tong, ``Dynamic knowledge graph
  alignment,'' in \emph{Proceedings of the AAAI Conference on Artificial
  Intelligence}, vol.~35, no.~5, 2021, pp. 4564--4572.

\bibitem{zhang2018variational}
Y.~Zhang, H.~Dai, Z.~Kozareva, A.~J. Smola, and L.~Song, ``Variational
  reasoning for question answering with knowledge graph,'' in \emph{AAAI},
  2018, pp. 6069--6076.

\bibitem{kipf2016semi}
T.~N. Kipf and M.~Welling, ``Semi-supervised classification with graph
  convolutional networks,'' in \emph{ICLR}, 2017, pp. 1--14.

\bibitem{velivckovic2018graph}
P.~Velickovic, G.~Cucurull, A.~Casanova, A.~Romero, P.~Li{\`{o}}, and
  Y.~Bengio, ``Graph attention networks,'' in \emph{ICLR}, 2018, pp. 1--12.

\bibitem{wang2018cross}
Z.~Wang, Q.~Lv, X.~Lan, and Y.~Zhang, ``Cross-lingual knowledge graph alignment
  via graph convolutional networks,'' in \emph{EMNLP}, 2018, pp. 349--357.

\bibitem{wu2019relation}
Y.~Wu, X.~Liu, Y.~Feng, Z.~Wang, R.~Yan, and D.~Zhao, ``Relation-aware entity
  alignment for heterogeneous knowledge graphs,'' in \emph{IJCAI}, 2019, pp.
  5278--5284.

\bibitem{cao2019multi}
Y.~Cao, Z.~Liu, C.~Li, Z.~Liu, J.~Li, and T.-S. Chua, ``Multi-channel graph
  neural network for entity alignment,'' in \emph{Proceedings of the 57th
  Annual Meeting of the Association for Computational Linguistics}, 2019, pp.
  1452--1461.

\bibitem{xu2019cross}
K.~Xu, L.~Wang, M.~Yu, Y.~Feng, Y.~Song, Z.~Wang, and D.~Yu, ``Cross-lingual
  knowledge graph alignment via graph matching neural network,'' in
  \emph{Proceedings of the 57th Annual Meeting of the Association for
  Computational Linguistics}, 2019, pp. 3156--3161.

\bibitem{TRUNG2020112883}
H.~T. Trung, N.~T. Toan, T.~Van~Vinh, H.~T. Dat, D.~C. Thang, N.~Q.~V. Hung,
  and A.~Sattar, ``A comparative study on network alignment techniques,''
  \emph{Expert Systems with Applications}, vol. 140, p. 112883, 2020.

\bibitem{huynh2019network}
T.~T. Huynh, C.~T. Duong, T.~H. Quyet, Q.~V.~H. Nguyen, A.~Sattar
  \emph{et~al.}, ``Network alignment by representation learning on structure
  and attribute,'' in \emph{Pacific Rim International Conference on Artificial
  Intelligence}, 2019, pp. 698--711.

\bibitem{sun2020knowledge}
Z.~Sun, C.~Wang, W.~Hu, M.~Chen, J.~Dai, W.~Zhang, and Y.~Qu, ``Knowledge graph
  alignment network with gated multi-hop neighborhood aggregation,'' in
  \emph{Proceedings of the AAAI Conference on Artificial Intelligence},
  vol.~34, no.~01, 2020, pp. 222--229.

\bibitem{zhao2020experimental}
X.~Zhao, W.~Zeng, J.~Tang, W.~Wang, and F.~Suchanek, ``An experimental study of
  state-of-the-art entity alignment approaches,'' \emph{IEEE Transactions on
  Knowledge \& Data Engineering}, no.~01, pp. 1--1, 2020.

\bibitem{nguyen2020entity}
T.~T. Nguyen, T.~T. Huynh, H.~Yin, V.~Van~Tong, D.~Sakong, B.~Zheng, and
  Q.~V.~H. Nguyen, ``Entity alignment for knowledge graphs with multi-order
  convolutional networks,'' \emph{IEEE Transactions on Knowledge and Data
  Engineering}, vol.~32, no.~13, pp. 1--14, 2021.

\bibitem{jumping_knowledge}
K.~Xu, C.~Li, Y.~Tian, T.~Sonobe, K.~Kawarabayashi, and S.~Jegelka,
  ``Representation learning on graphs with jumping knowledge networks,'' in
  \emph{International Conference on Machine Learning}, 2018, pp. 5449--5458.

\bibitem{MTransE}
M.~Chen, Y.~Tian, M.~Yang, and C.~Zaniolo, ``Multilingual knowledge graph
  embeddings for cross-lingual knowledge alignment,'' in \emph{IJCAI}, 2017,
  pp. 1511--1517.

\bibitem{el2015alex}
A.~El-Roby and A.~Aboulnaga, ``Alex: Automatic link exploration in linked
  data,'' in \emph{SIGMOD}, 2015, pp. 1839--1853.

\bibitem{chen2014unified}
M.~Chen, I.~W. Tsang, M.~Tan, and T.~J. Cham, ``A unified feature selection
  framework for graph embedding on high dimensional data,'' \emph{IEEE
  Transactions on Knowledge and Data Engineering}, vol.~27, no.~6, pp.
  1465--1477, 2014.

\bibitem{chen2019exploiting}
H.~Chen, H.~Yin, T.~Chen, Q.~V.~H. Nguyen, W.-C. Peng, and X.~Li, ``Exploiting
  centrality information with graph convolutions for network representation
  learning,'' in \emph{IEEE 35th International Conference on Data Engineering
  (ICDE)}, 2019, pp. 590--601.

\bibitem{JAPE}
Z.~Sun, W.~Hu, and C.~Li, ``Cross-lingual entity alignment via joint
  attribute-preserving embedding,'' in \emph{International Semantic Web
  Conference}, 2017, pp. 628--644.

\bibitem{zhu2017iterative}
H.~Zhu, R.~Xie, Z.~Liu, and M.~Sun, ``Iterative entity alignment via joint
  knowledge embeddings.'' in \emph{Proceedings of the Twenty-Sixth
  International Joint Conference on Artificial Intelligence}, 2017, pp.
  4258--4264.

\bibitem{sun2018bootstrapping}
Z.~Sun, W.~Hu, Q.~Zhang, and Y.~Qu, ``Bootstrapping entity alignment with
  knowledge graph embedding,'' in \emph{Proceedings of the 27th International
  Joint Conference on Artificial Intelligence}, 2018, pp. 4396--4402.

\bibitem{bordes2013translating}
A.~Bordes, N.~Usunier, A.~Garcia-Dur\'{a}n, J.~Weston, and O.~Yakhnenko,
  ``Translating embeddings for modeling multi-relational data,'' in
  \emph{Proceedings of the 26th International Conference on Neural Information
  Processing Systems}, 2013, pp. 2787--2795.

\bibitem{chen2018co}
M.~Chen, Y.~Tian, K.-W. Chang, S.~Skiena, and C.~Zaniolo, ``Co-training
  embeddings of knowledge graphs and entity descriptions for cross-lingual
  entity alignment,'' \emph{arXiv preprint arXiv:1806.06478}, 2018.

\bibitem{zhang2019multi}
Q.~Zhang, Z.~Sun, W.~Hu, M.~Chen, L.~Guo, and Y.~Qu, ``Multi-view knowledge
  graph embedding for entity alignment,'' \emph{IJCAI}, 2019.

\bibitem{pei2020rea}
S.~Pei, L.~Yu, G.~Yu, and X.~Zhang, ``Rea: Robust cross-lingual entity
  alignment between knowledge graphs,'' in \emph{KDD}, 2020, pp. 2175--2184.

\bibitem{chen2021passleaf}
Z.-M. Chen, M.-Y. Yeh, and T.-W. Kuo, ``Passleaf: A pool-based semi-supervised
  learning framework for uncertain knowledge graph embedding,'' in \emph{AAAI},
  vol.~35, no.~5, 2021, pp. 4019--4026.

\bibitem{wang2014knowledge}
Z.~Wang, J.~Zhang, J.~Feng, and Z.~Chen, ``Knowledge graph embedding by
  translating on hyperplanes,'' in \emph{Proceedings of the AAAI Conference on
  Artificial Intelligence}, vol.~28, no.~1, 2014.

\bibitem{dettmers2018convolutional}
T.~Dettmers, P.~Minervini, P.~Stenetorp, and S.~Riedel, ``Convolutional 2d
  knowledge graph embeddings,'' in \emph{Thirty-second AAAI conference on
  artificial intelligence}, 2018, pp. 1811--1818.

\bibitem{nguyen2018novel}
D.~Q. Nguyen, T.~D. Nguyen, D.~Q. Nguyen, and D.~Q. Phung, ``A novel embedding
  model for knowledge base completion based on convolutional neural network,''
  pp. 327--333, 2018.

\bibitem{liu2017analogical}
H.~Liu, Y.~Wu, and Y.~Yang, ``Analogical inference for multi-relational
  embeddings,'' in \emph{International conference on machine learning}.\hskip
  1em plus 0.5em minus 0.4em\relax PMLR, 2017, pp. 2168--2178.

\bibitem{shang2019end}
C.~Shang, Y.~Tang, J.~Huang, J.~Bi, X.~He, and B.~Zhou, ``End-to-end
  structure-aware convolutional networks for knowledge base completion,'' in
  \emph{AAAI}, vol.~33, no.~01, 2019, pp. 3060--3067.

\bibitem{sun2020re}
Z.~Sun, S.~Vashishth, S.~Sanyal, P.~Talukdar, and Y.~Yang, ``A re-evaluation of
  knowledge graph completion methods,'' \emph{ACL}, 2020.

\bibitem{brody2021attentive}
S.~Brody, U.~Alon, and E.~Yahav, ``How attentive are graph attention
  networks?'' \emph{arXiv preprint arXiv:2105.14491}, 2021.

\bibitem{greedymatch}
G.~Kollias, S.~Mohammadi, and A.~Grama, ``Network similarity decomposition
  (nsd): A fast and scalable approach to network alignment,'' \emph{IEEE
  Transactions on Knowledge and Data Engineering}, vol.~24, no.~12, pp.
  2232--2243, 2011.

\bibitem{mikolov2018advances}
T.~Mikolov, E.~Grave, P.~Bojanowski, C.~Puhrsch, and A.~Joulin, ``Advances in
  pre-training distributed word representations,'' in \emph{Proceedings of the
  International Conference on Language Resources and Evaluation (LREC 2018)},
  2018.

\bibitem{lehmann2015dbpedia}
J.~Lehmann, R.~Isele, M.~Jakob, A.~Jentzsch, D.~Kontokostas, P.~N. Mendes,
  S.~Hellmann, M.~Morsey, P.~Van~Kleef, S.~Auer \emph{et~al.}, ``Dbpedia--a
  large-scale, multilingual knowledge base extracted from wikipedia,''
  \emph{Semantic web}, vol.~6, no.~2, pp. 167--195, 2015.

\bibitem{suchanek2008yago}
F.~M. Suchanek, G.~Kasneci, and G.~Weikum, ``Yago: A large ontology from
  wikipedia and wordnet,'' \emph{Journal of Web Semantics}, vol.~6, no.~3, pp.
  203--217, 2008.

\bibitem{sun2017cross}
Z.~Sun, W.~Hu, and C.~Li, ``Cross-lingual entity alignment via joint
  attribute-preserving embedding,'' in \emph{International Semantic Web
  Conference}.\hskip 1em plus 0.5em minus 0.4em\relax Springer, 2017, pp.
  628--644.

\bibitem{userIdentityBenchmark}
K.~Shu, S.~Wang, J.~Tang, R.~Zafarani, and H.~Liu, ``User identity linkage
  across online social networks: A review,'' \emph{Acm Sigkdd Explorations
  Newsletter}, vol.~18, no.~2, pp. 5--17, 2017.

\bibitem{chen2019rgcn}
J.~Chen, H.~Hou, J.~Gao, Y.~Ji, and T.~Bai, ``Rgcn: Recurrent graph
  convolutional networks for target-dependent sentiment analysis,'' in
  \emph{International Conference on Knowledge Science, Engineering and
  Management}.\hskip 1em plus 0.5em minus 0.4em\relax Springer, 2019, pp.
  667--675.

\bibitem{vashishth2019composition}
S.~Vashishth, S.~Sanyal, V.~Nitin, and P.~Talukdar, ``Composition-based
  multi-relational graph convolutional networks,'' in \emph{International
  Conference on Learning Representations}, 2019, pp. 1--14.

\bibitem{qiao2021context}
Z.~Qiao, Z.~Ning, Y.~Du, and Y.~Zhou, ``Context-enhanced entity and relation
  embedding for knowledge graph completion (student abstract),'' in
  \emph{Proceedings of the AAAI Conference on Artificial Intelligence},
  vol.~35, no.~18, 2021, pp. 15\,871--15\,872.

\bibitem{che2020parame}
F.~Che, D.~Zhang, J.~Tao, M.~Niu, and B.~Zhao, ``Parame: Regarding neural
  network parameters as relation embeddings for knowledge graph completion,''
  in \emph{Proceedings of the AAAI Conference on Artificial Intelligence},
  vol.~34, no.~03, 2020, pp. 2774--2781.

\bibitem{zhang2020few}
C.~Zhang, H.~Yao, C.~Huang, M.~Jiang, Z.~Li, and N.~V. Chawla, ``Few-shot
  knowledge graph completion,'' in \emph{Proceedings of the AAAI Conference on
  Artificial Intelligence}, vol.~34, no.~03, 2020, pp. 3041--3048.

\bibitem{song2021robust}
L.~Song, J.~Wu, M.~Yang, Q.~Zhang, Y.~Li, and J.~Yuan, ``Robust knowledge
  transfer via hybrid forward on the teacher-student model,'' in \emph{AAAI},
  vol.~35, no.~3, 2021, pp. 2558--2566.

\bibitem{tseng2021toward}
W.-C. Tseng, J.-S. Lin, Y.-M. Feng, and M.~Sun, ``Toward robust long range
  policy transfer,'' in \emph{Proceedings of the AAAI Conference on Artificial
  Intelligence}, vol.~35, no.~11, 2021, pp. 9958--9966.

\end{thebibliography}


\end{document}